\documentclass[10pt,twocolumn,letterpaper]{article}

\usepackage{cvpr}              

\usepackage{caption}

\definecolor{cvprblue}{rgb}{0.21,0.49,0.74}
\usepackage[pagebackref,breaklinks,colorlinks,allcolors=cvprblue]{hyperref}

\def\paperID{*****} 
\def\confName{CVPR}
\def\confYear{2026}

\title{PanoWorld: A Generative Spatial World Model for Consistent Whole-House Panorama Synthesis}

\author{
Jinrang Jia \quad Zhenjia Li \quad Yijiang Hu \quad Yifeng Shi\\
Ke Holdings Inc.\\
Beijing, China\\
}

\begin{document}

\twocolumn[{
\renewcommand\twocolumn[1][]{#1}
\maketitle

\begin{center}
  \includegraphics[width=\textwidth]{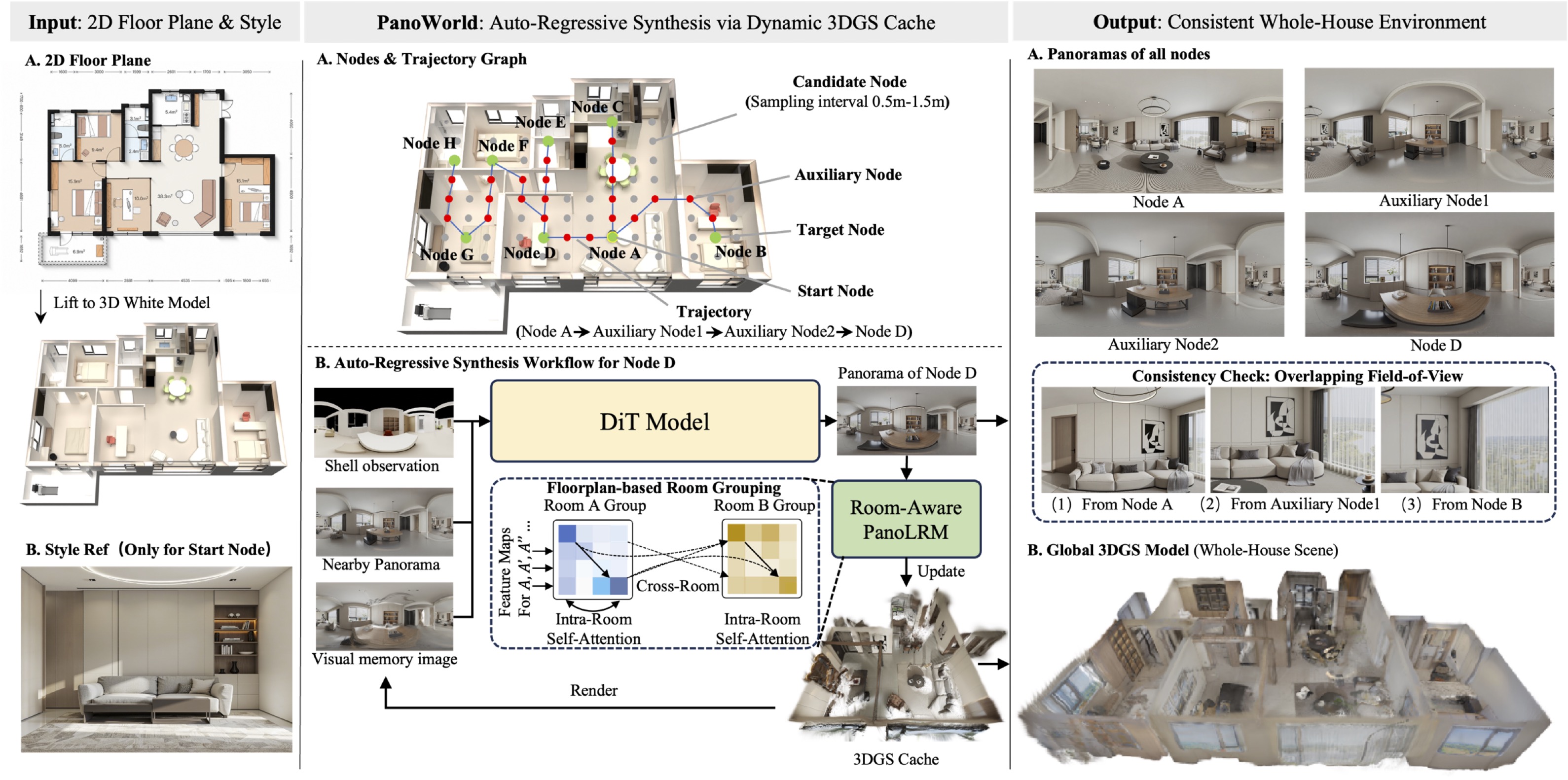}
  \captionof{figure}{\textbf{Teaser of PanoWorld.} Given a floorplan and a style reference, PanoWorld synthesizes a node-based whole-house panorama tour. A floorplan-derived geometric proxy anchors the global structure, while a dynamic 3DGS cache progressively expands along the navigation path and provides renderable spatial memory. The generated panoramas preserve photorealistic detail and cross-room consistency, e.g., doorway geometry and material appearance remain aligned when viewing the bedroom from the living room and the living room from the bedroom.}
  \label{fig:teaser}
\end{center}

\vspace{1em}
}]

\begin{abstract}
Generating a consistent whole-house VR tour from a floorplan and style reference requires both photorealistic panoramas and cross-view spatial coherence. Pure 2D generators produce appealing single panoramas but re-imagine geometry and materials when the viewpoint changes, whereas monolithic 3D generation becomes expensive and loses fine texture at multi-room scale. We introduce PanoWorld, a generative spatial world model that treats whole-house synthesis as autoregressive generation of node-based 360-degree panoramas, matching the discrete navigation used by real VR tour products. PanoWorld uses a floorplan-derived 3D shell as a global geometric proxy and a dynamic 3D Gaussian Splatting cache as renderable spatial memory. A feed-forward panoramic LRM designed for metric-scale multi-room 360-degree inputs lifts generated panoramas into local 3DGS updates, while Room-aware Group Attention suppresses cross-room feature interference. A topology-aware progressive caching strategy fuses these local updates without repeatedly reconstructing the full history. By decoupling shell-based geometry guidance from cache-rendered visual memory, PanoWorld preserves high-frequency 2D synthesis quality while improving cross-node layout and material consistency.  The project link is https://jjrcn.github.io/PanoWorld-project-home/
\end{abstract}    
\section{Introduction}
\label{sec:intro}
Synthesizing immersive, multi-room indoor environments from sparse architectural inputs remains a persistent challenge in spatial generation. Its difficulty goes far beyond single-view realism: a whole-house tour spans multiple rooms, doorways, corridors, and long-range visibility, requiring overlapping regions across viewpoints to preserve geometry, furniture layout, material identity, and fine details simultaneously.

Existing generation paradigms struggle to satisfy these requirements simultaneously. 2D diffusion models \cite{feng2023diffusion360seamless360degree, Zhang2024PanFusion, Ni_2025_ICCV} can synthesize visually rich panoramas with realistic lighting and high-frequency texture, but they usually lack persistent spatial memory. As the camera moves, the same doorway, wall, or sofa may be regenerated with a different shape, position, or material. Global 3D representations such as NeRF \cite{mildenhall2020nerf}, 3DGS \cite{kerbl3Dgaussians, jia2026gaussianoncecontrollable3d, zhang2024pansplat}, or mesh-based scenes \cite{fridman2023scenescape, hollein2023text2room, vidanapathirana2021plan2scene} provide a more natural route to consistency, yet directly generating a single detailed multi-room asset is costly. At house scale, these methods often face high memory usage, slow inference, and a loss of the texture fidelity that makes 2D generative models attractive for commercial visualization.

Our approach is motivated by the operational logic of commercial VR tours: they are predominantly node-based rather than continuous 6-DoF environments. Users stand at one panorama node, inspect the scene, and jump to another nearby node. This suggests a different formulation. Instead of forcing a monolithic 3D model to be high-quality everywhere, we can generate a set of high-resolution panorama nodes that are directly deliverable, while using a lightweight renderable 3D memory to make the nodes agree with each other.

We propose PanoWorld, a generative spatial world model for consistent whole-house panorama synthesis. PanoWorld first converts the floorplan into a coarse 3D shell that provides a global coordinate frame, room boundaries, doorway connectivity, and viewpoint visibility. The shell is not the final visual asset; it is rendered at target and auxiliary viewpoints to provide geometric guidance. Starting from an initial node, PanoWorld synthesizes a furnished panorama conditioned on the shell-derived proxy and the style reference, then lifts it into an initial 3DGS cache. For each subsequent node, the system renders visual memory from the current cache, combines it with the geometric proxy and nearby panoramas, generates the next panorama, and writes the new observation back into the cache.

Two components make this autoregressive loop scalable to multi-room scenes. First, we design a feed-forward panoramic LRM for metric-scale, multi-room 360-degree inputs. To our knowledge, this is the first LRM-style module aimed at whole-house multi-room reconstruction from multi-view panoramas in a single feed-forward pass. To avoid mixing unrelated evidence across walls, the model uses Room-aware Group Attention: panoramas interact densely within the same room, while doorway or boundary nodes provide restricted communication between connected rooms. Second, we introduce Topology-aware Progressive 3DGS Caching. Rather than feeding all historical panoramas into the LRM after every step, PanoWorld updates the cache using the new node, same-room history, and adjacent boundary nodes, then fuses local Gaussians into the global cache through alignment, confidence fusion, and visibility pruning. This keeps the spatial memory growing with the tour while avoiding full-history reconstruction.

Finally, PanoWorld decouples geometric and appearance guidance. The floorplan shell constrains walls, openings, floors, ceilings, and large-scale layout, while the 3DGS cache preserves colors, materials, and high-frequency details in overlapping views. This separation lets the 2D generator retain photorealistic texture quality without losing cross-node consistency. In summary, our contributions are: (1) a node-based world-model formulation for whole-house VR panorama synthesis; (2) a room-aware panoramic LRM for metric-scale whole-house multi-room panorama reconstruction, with masked attention to suppress cross-room feature interference; (3) a topology-aware progressive 3DGS cache for scalable spatial memory; and (4) a decoupled conditioning strategy that improves layout and material consistency across dense panorama nodes.

\section{Related Work}
\label{sec:related}
\subsection{Text/Image-to-Panorama Generation}

Recent diffusion models have substantially advanced 360-degree panoramic image synthesis \cite{chen2022text2light,Ni_2025_ICCV}. Representative systems address panorama outpainting, correspondence-aware multi-view generation, recursive environment expansion, and projection-aware text-to-360 synthesis \cite{Wu2023PanoDiffusion,Tang2023MVDiffusion,Li2023PanoGen,Zhang2024PanFusion}. These methods improve single-node quality and seam consistency, but mainly target one panorama or a synchronized view set. PanoWorld instead targets whole-house tours, where many panorama nodes must remain consistent along long paths across rooms and doorways.

\subsection{Large Reconstruction Models}

Large Reconstruction Models have shown that feed-forward networks can rapidly lift images into 3D representations. LRM predicts an object-level NeRF from a single image using a large transformer trained on multi-view data \cite{Hong2024LRM}. Instant3D combines sparse-view generation with a transformer-based reconstructor for fast text-to-3D assets \cite{Li2024Instant3D}, and TripoSR improves single-image reconstruction speed and mesh quality \cite{Tochilkin2024TripoSR}. More recent models extend this direction with multi-view or Gaussian representations, such as pixelSplat \cite{charatan2024pixelsplat}, GS-LRM \cite{he2024gslrm}, LGM \cite{Tang2024LGM}, and M-LRM \cite{Li2024MLRM}. However, most LRM-style systems are designed for objects or compact scenes, where all input views describe a shared target. Whole-house panoramas introduce room-level topology: views from different rooms may be geometrically disconnected by walls and should not freely attend to each other. To our knowledge, existing LRM-style systems have not targeted metric-scale whole-house, multi-room reconstruction from multi-view panoramas in a single feed-forward pass. PanoWorld addresses this gap with a room-aware panoramic LRM and topology-aware local updates.

\subsection{Indoor Layout and Floorplan-Conditioned Synthesis}

Indoor scene synthesis extensively utilizes structural priors like scene graphs or floorplans. Graph-to-3D \cite{dhamo2021graph} leverages scene graphs for 3D object arrangement, while Plan2Scene \cite{vidanapathirana2021plan2scene} converts floorplans and photos into textured 3D meshes. For interior layouts, transformer-based models like ATISS \cite{Paschalidou2021ATISS} and SceneFormer \cite{wang2021sceneformer} autoregressively generate plausible furniture arrangements. Recently, diffusion models have advanced this domain: HouseDiffusion \cite{shabani2023housediffusion} and DiffuScene \cite{tang2024diffuscene} generative model vector floorplans and 3D layouts, and MiDiffusion \cite{Hu2024MiDiffusion} formulates floor-conditioned synthesis via mixed discrete-continuous diffusion. While these methods focus on structural modeling or object arrangement, PanoWorld instead uses the floorplan as a global geometric proxy for photorealistic panorama generation, coupling layout constraints with dynamic 3DGS memory to ensure cross-view appearance consistency.
\section{Method}
\label{sec:method}

\subsection{Problem Formulation and Overview}

Given a 2D floorplan \(F\), a style condition \(s\), and a set of target panorama poses \(\mathcal{V}^{tar}=\{v_i\}_{i=1}^{N}\), PanoWorld generates a set of furnished 360-degree panoramas \(\mathcal{I}^{tar}=\{I_i\}_{i=1}^{N}\) and maintains a renderable 3DGS cache \(\mathcal{C}\) as spatial memory. The target poses and auxiliary poses form a topological node graph \(\mathcal{G}=(\mathcal{V},\mathcal{E})\), where nodes are camera poses and edges indicate navigation adjacency. The output is optimized for node-based VR tours: the panoramas are the primary deliverable, while \(\mathcal{C}\) provides memory and guidance rather than serving as a perfect continuous 6-DoF asset.

PanoWorld follows an autoregressive loop. First, the floorplan is converted into a coarse 3D shell and rendered at each node to obtain a geometric proxy. The starting panorama is synthesized from this proxy and the style condition, then lifted by a panoramic LRM into an initial 3DGS cache. For each subsequent node, the system renders visual memory from the current cache, combines it with the geometric proxy and nearby panoramas, synthesizes the next high-resolution panorama, and updates the cache with a local 3DGS increment. The crux of this autoregressive formulation lies in ensuring scalable, room-aware consistency: concurrent views within the same room must reinforce underlying geometry, whereas views separated by walls should not freely exchange appearance evidence.

\subsection{Global Geometric Proxy from Floorplan}

The floorplan-derived geometry is used as a structural interface, not as a central contribution of this work. We assume an off-the-shelf or engineering pipeline converts \(F\) into a coarse 3D shell \(\mathcal{S}\) containing walls, floors, ceilings, room labels, and doorway connectivity. For a node \(v_i\), we render a shell observation \(B_i=R_{\mathcal{S}}(v_i)\), then convert it into a compact geometric proxy \(G_i\), including normal and semantic segmentation maps. This proxy provides stable low-frequency constraints for wall layout, openings, and room extent. It deliberately contains no final texture, allowing the 2D generator to synthesize photorealistic appearance while respecting the global structure.

\subsection{Topology-Guided Node and Path Sampling}

PanoWorld uses the floorplan topology to organize generation order. We choose a starting node with high graph centrality or low average path cost to the target nodes, then connect target poses through room adjacency and doorway constraints. When two adjacent targets are far apart, auxiliary nodes are inserted so that neighboring viewpoints have sufficient visual overlap; in our implementation this spacing is typically 0.5--1.5m. These auxiliary nodes are not necessarily part of the final user-facing tour, but they make the autoregressive process smoother and provide intermediate observations for cache growth. Since path planning is not the focus of this paper, we use this module as a simple deterministic scaffold for the subsequent room-aware generation.

\begin{figure}[t]
  \centering
  \includegraphics[width=\linewidth]{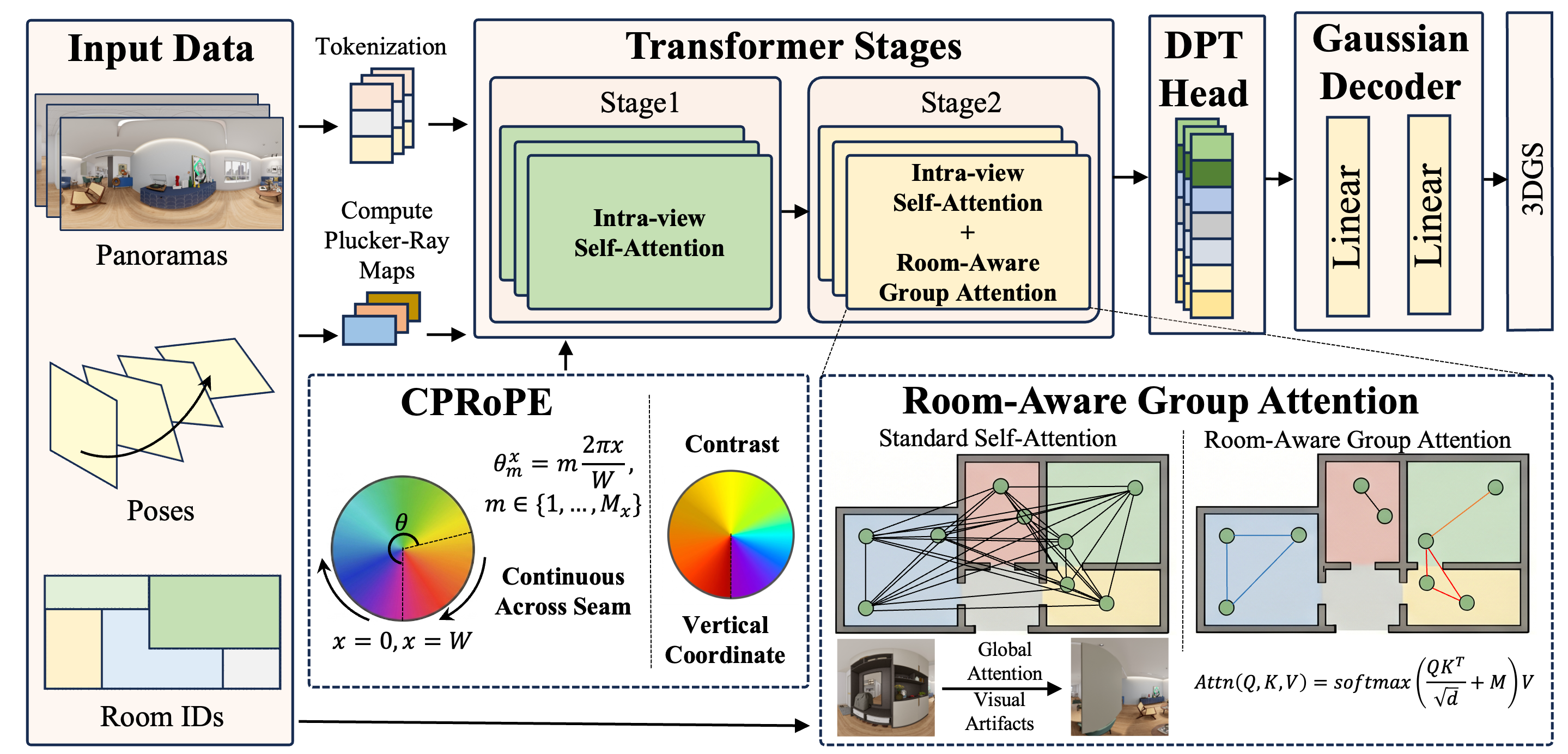}
  \caption{\textbf{Room-aware panoramic LRM.} Grouped attention allows dense intra-room interaction and restricted cross-room communication only through topological boundaries.}
  \label{fig:room_aware_lrm}
\end{figure}

\subsection{Room-Aware Panoramic LRM}

The panoramic LRM is designed for metric-scale whole-house reconstruction from multi-view 360-degree observations in a single feed-forward pass. In PanoWorld's progressive loop, it is applied to topology-selected contexts so that the same model predicts local 3DGS updates without reconstructing the entire history at every node. Given a local context set \(\mathcal{H}_t\) of generated panoramas, poses, geometric proxies, and room labels, the model predicts Gaussian primitives \(\Delta\mathcal{C}_t=\{(\mu_k,q_k,\sigma_k,\alpha_k,c_k)\}_k\), where \(\mu_k\) is the 3D mean, \(q_k\) the rotation, \(\sigma_k\) the anisotropic scale, \(\alpha_k\) the opacity, and \(c_k\) the color feature. Each panorama is encoded with an equirectangular image encoder, and the decoder maps fused tokens to Gaussian parameters in the global coordinate frame.

\subsubsection{Panoramic Position Encoding}
\label{cprope}
We adapt the Plucker-ray and PRoPE encoding used in multi-view reconstruction models \cite{li2025cameras,kang2025multi} to equirectangular panoramas with two changes. First, since a panorama has no single pinhole intrinsic matrix, we replace Plucker rays built from extrinsics and intrinsics with extrinsics-only Plucker rays. For a token at \((x,y)\), we obtain its spherical unit direction \(r(x,y)\), transform it by the camera rotation \(R_i\), and form
\begin{equation}
\rho_{i,x,y}=(d_{i,x,y}, o_i \times d_{i,x,y}),\quad d_{i,x,y}=R_i r(x,y),
\end{equation}
where \(o_i\) is the camera center. Second, because the left and right boundaries of a panorama are adjacent, we replace the horizontal PRoPE coordinate with a periodic one. Let \(x\in\{0,\ldots,W-1\}\) be the horizontal token index and \(W\) be the number of horizontal panorama tokens. We first map \(x\) to an angular coordinate
\begin{equation}
\phi(x)=\frac{2\pi x}{W}.
\end{equation}
If \(M_x\) sine-cosine frequency pairs are allocated to the horizontal RoPE branch, the \(m\)-th pair uses the integer-harmonic phase
\begin{equation}
\theta^x_m(x)=m\,\phi(x),\quad m\in\{1,\ldots,M_x\},
\end{equation}
and precomputes \((\cos\theta^x_m(x),\sin\theta^x_m(x))\). Here \(m\) indexes the horizontal frequency pair, not an image location, and \(M_x\) is the number of such pairs, i.e., half of the feature dimension assigned to this horizontal branch. A virtual position \(x=W\) therefore has the same coefficients as \(x=0\), since the phase differs by \(2\pi m\). Unlike the non-circular vertical branch, which keeps the standard RoPE frequency schedule, the circular horizontal branch uses integer harmonics so that every frequency is periodic over the panorama width. This Circular PRoPE (CPRoPE) keeps the geometric camera encoding of PRoPE while making attention continuous across the panorama seam.

\subsubsection{Room-Aware Group Attention}
Standard self-attention is poorly matched to multi-room panoramas. If all view tokens attend globally, texture from one room can leak through walls into another room, producing ghosted geometry or duplicated materials. We therefore introduce Room-aware Group Attention. For tokens from nodes \(i\) and \(j\), attention is allowed when the nodes belong to the same room or when they correspond to topologically connected doorway/boundary nodes. Otherwise, the attention logit is masked:
\begin{equation}
\mathrm{Attn}(Q,K,V)=\mathrm{softmax}\left(\frac{QK^\top}{\sqrt{d}}+M\right)V,
\end{equation}
where \(M_{ij}=0\) for valid same-room or doorway-connected pairs and \(M_{ij}=-\infty\) for unrelated cross-room pairs. This mask preserves dense interaction within a room while permitting controlled information exchange across actual openings. As a result, the LRM can aggregate redundant observations of the same space without confusing visually similar but physically separated regions.

\subsubsection{Training Objective}
The panoramic LRM is trained as a feed-forward memory extractor. Predicted Gaussians are rendered back to held-out panorama views and supervised by an image L2 loss, a VGG19 perceptual loss, an opacity regularizer, and a depth loss on the Gaussian positions induced by input pixels. Importantly, the depth term does not supervise the rendered depth map. Instead, for valid input pixels \(\Omega\), it compares the camera-space depth \(\hat d_p\) of the predicted Gaussian position with the corresponding target depth \(d_p\). We use a log-depth L1 term and a scale-invariant log term:
\begin{equation}
\mathcal{L}_{\mathrm{log}}=\frac{1}{|\Omega|}\sum_{p\in\Omega}\left|\log(\hat d_p+1)-\log(d_p+1)\right|,
\end{equation}
\begin{equation}
\begin{aligned}
\delta_p &= \log(\hat d_p+\epsilon)-\log(d_p+\epsilon),\\
\mathcal{L}_{\mathrm{si}}
&=0.1\sqrt{
\frac{1}{|\Omega|}\sum_{p}\delta_p^2
-0.85\left(\frac{1}{|\Omega|}\sum_{p}\delta_p\right)^2
+\epsilon } .
\end{aligned}
\end{equation}
The depth loss is \(\mathcal{L}_{\mathrm{depth}}=\mathcal{L}_{\mathrm{log}}+\mathcal{L}_{\mathrm{si}}\), and the total objective is
\begin{equation}
\mathcal{L}=\lambda_{2}\mathcal{L}_{2}+\lambda_{\mathrm{perc}}\mathcal{L}_{\mathrm{perc}}+\lambda_{\alpha}\mathcal{L}_{\alpha}+\lambda_{d}\mathcal{L}_{\mathrm{depth}},
\end{equation}
where the weights are set to $\lambda_2 = 1.0$, $\lambda_{\mathrm{perc}} = 0.1$, $\lambda_{\alpha} = 0.05$, and $\lambda_{d} = 0.5$. This training objective encourages the cache to be geometrically useful for future view guidance rather than merely producing a plausible standalone reconstruction.

\subsection{Topology-Aware Progressive 3DGS Caching}

A naive autoregressive system could rerun the LRM on all previously generated panoramas after every new node. This quickly becomes impractical: memory and attention cost grow with path length, and distant rooms repeatedly consume computation even when they are irrelevant to the current viewpoint. PanoWorld instead maintains a dynamic cache \(\mathcal{C}_t\) and updates it locally. For a new node \(v_t\), we construct a fixed-size context
\begin{equation}
\mathcal{H}_t=\{v_t\}\cup \mathcal{N}_{same}(v_t)\cup \mathcal{N}_{door}(v_t),
\end{equation}
where \(\mathcal{N}_{same}\) contains nearby generated nodes in the same room and \(\mathcal{N}_{door}\) contains boundary nodes connected through doorways. The room-aware LRM predicts only a local update \(\Delta\mathcal{C}_t\), which is then merged into the global cache.

\begin{figure}[t]
  \centering
  \includegraphics[width=\linewidth]{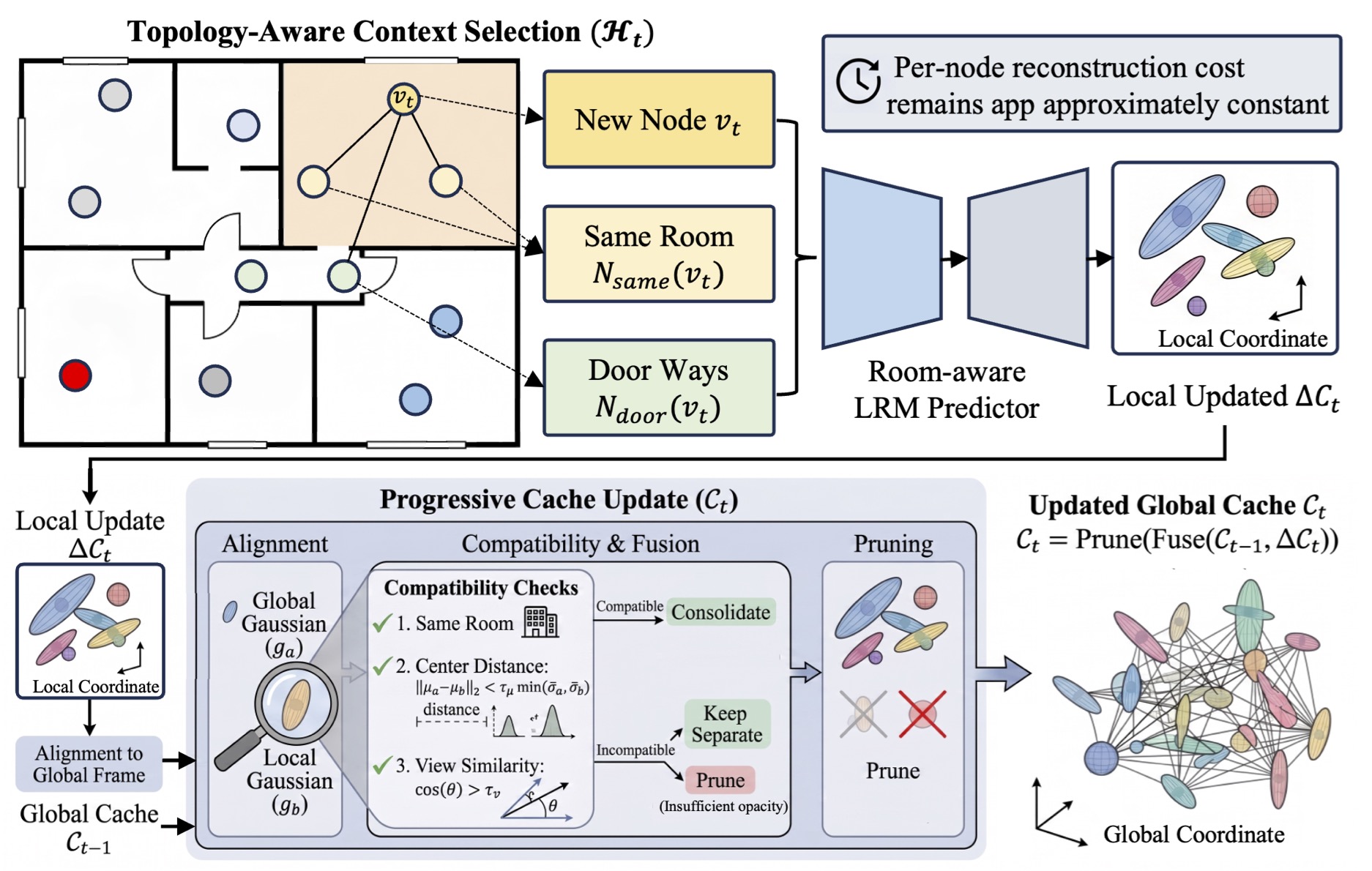}
  \caption{\textbf{Progressive 3DGS caching.} PanoWorld updates spatial memory through local topology-aware increments instead of full-history reconstruction.}
  \label{fig:progressive_cache}
\end{figure}

\subsubsection{Progressive Cache Update}
The merge step is deliberately conservative. Alignment transforms local Gaussians into the global coordinate frame using the known panorama poses and shell coordinate system. We only mark a new Gaussian and an existing Gaussian as compatible when they belong to the same room, their centers satisfy $\|\mu_a-\mu_b\|_2 < \tau_\mu \min(\bar\sigma_a,\bar\sigma_b)$, and their supporting viewing directions have a cosine similarity larger than $\tau_v$, where $\bar\sigma$ denotes the mean Gaussian scale. Compatible Gaussians are merged, while incompatible primitives are kept separate or pruned if their opacity is insufficient.

We avoid aggressive rule-based color averaging (e.g., computing the arithmetic mean of Spherical Harmonics (SH) coefficients across all bands) because such numerical blending destroys high-frequency view-dependent features, irreversibly making 3DGS renderings blurry and locally inconsistent. Instead, we adopt a confidence-based feature selection strategy during consolidation. The geometric properties (position and covariance) of the fused Gaussian are derived via an opacity-weighted average of the original primitives. For appearance attributes, we smoothly blend only the zero-order SH coefficients representing the base color. Conversely, the higher-order SH coefficients are strictly inherited from the dominant Gaussian—the one with higher opacity under the current supporting view—thereby maximally preserving local structural sharpness.

In PanoWorld, the cache serves as a spatial memory rather than the final appearance source; residual inconsistencies are handled by the subsequent 2D generator, which also leverages the nearby original panorama as a strong appearance-consistency reference. The resulting cache update is defined as:
\begin{equation}
\mathcal{C}_t=\mathrm{Prune}\left(\mathrm{Fuse}(\mathcal{C}_{t-1},\Delta\mathcal{C}_t)\right).
\end{equation}
Because the context size is bounded by local topology rather than the full generation history, the per-node reconstruction cost remains approximately constant. At the same time, the cache still grows into a whole-house memory that can be rendered from future nodes to enforce appearance continuity.

\subsubsection{Cross-Room Memory Filtering}
When rendering the cache from a new room, previously reconstructed Gaussians may represent the front side of a wall in the old room but become visible as incorrect back-side texture from the new room. We filter these large erroneous memory regions using the floorplan shell depth. Let \(D_{\mathcal{C}}(u)\) be the cache-rendered depth at pixel \(u\) and \(D_{\mathcal{S}}(u)\) the shell-rendered depth. If
\begin{equation}
D_{\mathcal{C}}(u) > D_{\mathcal{S}}(u)+\tau_D,
\end{equation}
the memory pixel is behind the first shell surface and is therefore marked invalid in the visual memory image by setting its value to 255. This simple depth gate prevents old-room wall textures from leaking into the new room before the 2D generator synthesizes the next panorama.

\begin{figure}[t]
  \centering
  \includegraphics[width=\linewidth]{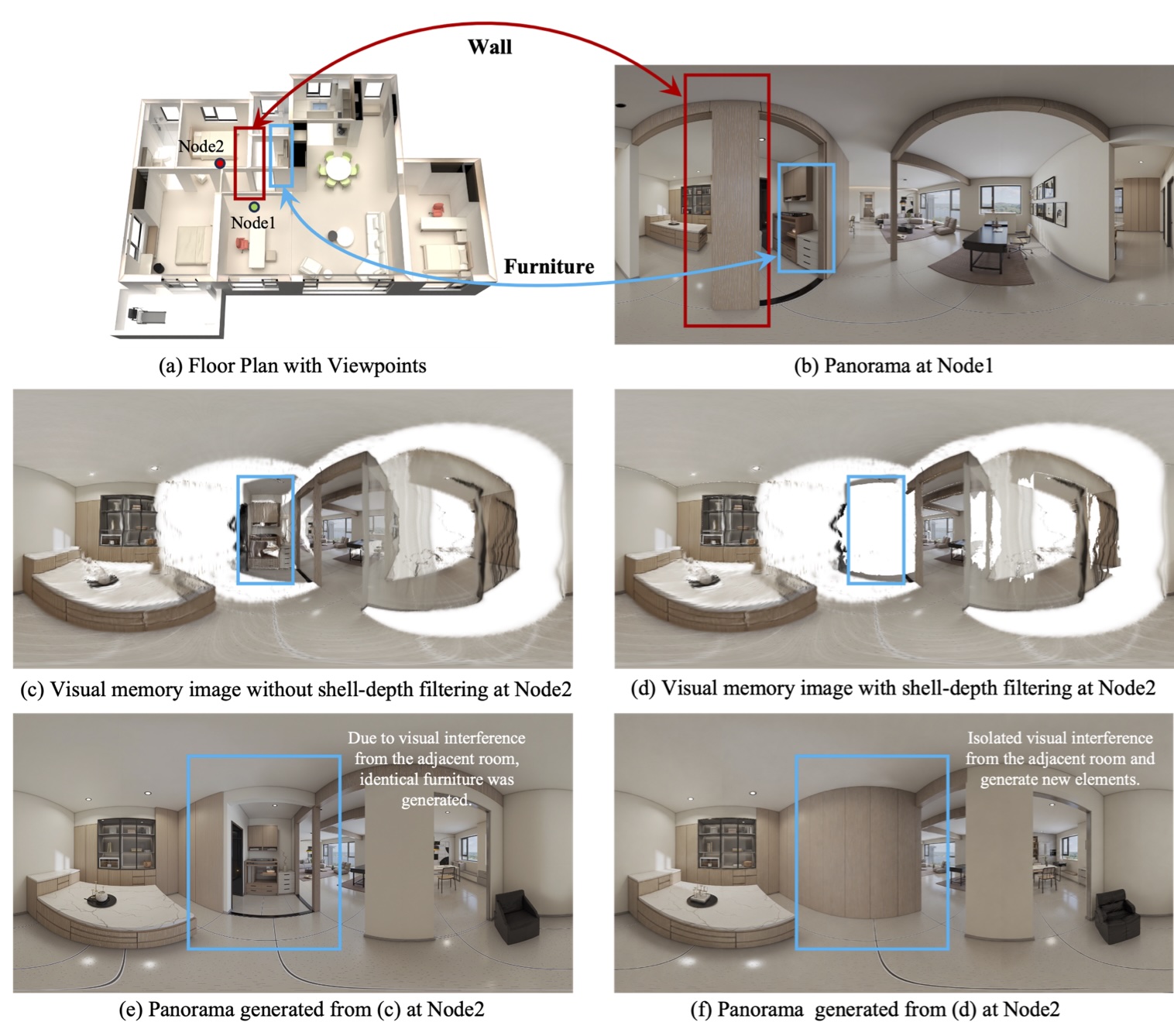}
  \caption{\textbf{Cross-room memory filtering.} Shell depth removes cache pixels that lie behind the first visible room surface and would otherwise introduce large erroneous textures.}
  \label{fig:cross_room_filtering}
\end{figure}

\subsection{Auto-Regressive Panorama Synthesis with Decoupled Guidance}

The 2D panorama generator uses Qwen-Image-Edit \cite{wu2025qwenimagetechnicalreport} as its backbone and is responsible for final visual fidelity. It also adopts the Plucker extrinsics-only rays with CPRoPE described in Sec.~\ref{cprope} to preserve panoramic wraparound continuity in its attention layers. For the starting node \(v_0\), it synthesizes \(I_0=\Phi(G_0,s)\) from the shell-derived geometry and style condition. The style condition is used only at this initialization step. For a later node \(v_t\), PanoWorld renders the current cache into the target pose and obtains a visual memory image \(V_t=R_{\mathcal{C}_{t-1}}(v_t)\). The generator then predicts
\begin{equation}
I_t=\Phi(G_t,V_t,I_{p(t)}),
\end{equation}
where \(G_t\) is the geometric proxy and \(I_{p(t)}\) is a nearby generated panorama. The nearby panorama provides local appearance context and carries the style forward, while the cache rendering supplies spatially aligned memory for regions that have already been observed.

The key design is to decouple geometry and appearance. The shell-derived proxy is injected as a structural condition, constraining walls, openings, floors, ceilings, and large-scale room layout. The cache-rendered memory is injected as an appearance condition, preserving colors, materials, and high-frequency details in overlapping regions. Treating these two sources separately prevents texture memory from overriding global geometry and prevents the coarse shell from suppressing photorealistic details. Invalid cache pixels, including those removed by the cross-room depth gate, are encoded directly in \(V_t\) and are ignored by the generator as missing memory. After \(I_t\) is generated, the panoramic LRM extracts \(\Delta\mathcal{C}_t\), the progressive cache is updated, and the loop proceeds to the next node. This gives PanoWorld a practical balance: high-quality 2D panoramas remain the final output, while 3DGS memory provides the cross-node discipline needed for coherent whole-house tours.

\section{Experiments}
\label{sec:experiments}
\subsection{Experimental Settings}

\subsubsection{Training Data}

We use three data sources. First, we render 6,813 3D-FRONT houses \cite{fu20213dfront3dfurnishedrooms} into approximately 200K panoramas with depth. Second, we use RealSee3D \cite{Li2025realsee3d_data}, containing 10K house scenes and 299,073 panoramas with depth. Third, we collect 2.5M private 2D panoramas without 3D annotations, which are used only to improve the visual quality of the 2D panorama generator. The 3D-FRONT and RealSee3D data are used to train both the panoramic LRM and the 2D generator, while the private 2D data are used only for the 2D generator. Representative training examples and room-level BEV maps are shown in the supplementary material.

\begin{table*}[t]
  \centering
  \caption{\textbf{Dataset summary.} 3D-FRONT and RealSee3D provide 3D/depth supervision for both the panoramic LRM and the 2D generator, while private 2D panoramas are used only to improve visual synthesis quality.}
  \label{tab:datasets}
  \begin{tabular}{llllll}
    \hline
    Split & Source & Scenes & Panoramas / nodes & 3D/depth & Usage \\
    \hline
    Train & 3D-FRONT rendered & 6,813 & about 200K & yes & LRM and 2D generator \\
    Train & RealSee3D & 10K & 299,073 & yes & LRM and 2D generator \\
    Train & Private 2D panoramas & - & 2.5M & no & 2D generator only \\
    Eval & Floorplan benchmark & 7 & 126 panoramas / 42 nodes & shell and depth & panorama synthesis \\
    Eval & RealSee3D held-out & 50 & 400 and 600 & yes & LRM reconstruction \\
    \hline
  \end{tabular}
\end{table*}

\begin{table}[t]
  \centering
  \caption{\textbf{Quantitative comparison on panorama synthesis.} HPSv3 measures single-node aesthetic quality, CLIP-I Style measures image-reference style consistency, and cross-node consistency is evaluated by Overlap PSNR (PSNR$_{\text{ov}}$).}
  \label{tab:panorama_quant}
  \begin{tabular}{lccc}
    \hline
    Method & HPSv3 $\uparrow$ & CLIP-I Style $\uparrow$ & PSNR$_{\text{ov}}$ $\uparrow$ \\
    \hline
    DreamHome-Pano & 8.5711 & 0.7785 & 15.4022 \\
    Pano2room & 2.1771 & 0.6796 & 15.7788 \\
    Nano Banana 2 & \textbf{9.5483} & \textbf{0.7940} & 14.7476 \\
    Seedream-4.5-Edit & 7.0733 & 0.7829 & 15.3870 \\
    OmniRoam & 6.1492 & 0.7201 & 16.3862 \\
    PanoWorld & 7.9564 & 0.7577 & \textbf{22.1365} \\
    \hline
  \end{tabular}
\end{table}

\begin{table}[t]
  \centering
  \caption{\textbf{Whole-house reconstruction quality on held-out RealSee3D scenes.} Metrics are computed from panorama renderings of reconstructed 3D representations.}
  \label{tab:lrm_reconstruction}
  \begin{tabular}{lcccc}
    \hline
    Method & Inputs & PSNR $\uparrow$ & SSIM $\uparrow$ & LPIPS $\downarrow$ \\
    \hline
    MVP & 8 & 21.0370 & 0.8145 & 0.3044 \\
    Adapt-Splat & 8 & 21.2418 & 0.8195 & 0.2978\\
    WorldMirror 2.0 & 8 & 13.3344 & 0.5402 & 0.5690 \\
    PanoWorld & 8 & \textbf{29.2361} & \textbf{0.8880} & \textbf{0.2225}\\
    \hline
    MVP & 12 & 20.8342 & 0.8090 & 0.3095 \\
    Adapt-Splat & 12 & 21.5156 & 0.8240 & 0.2906 \\
    WorldMirror 2.0 & 12 & 12.7541 & 0.5160 & 0.5943 \\
    PanoWorld & 12 & \textbf{28.8003} & \textbf{0.8817} & \textbf{0.2299} \\
    \hline
  \end{tabular}
\end{table}

\subsubsection{Evaluation Data}

For panorama synthesis, we construct and will release an evaluation dataset based on private data. It contains seven representative real floorplans, the corresponding 3D shell assets, and three style settings for each floorplan. For each sampled viewpoint, we provide shell-rendered placeholder images and depth maps. Across these floorplans, we sample 42 panorama viewpoints, yielding 126 evaluation panoramas under the three style conditions for evaluating image quality, style consistency, and cross-node consistency. For whole-house LRM reconstruction, we hold out 50 RealSee3D scenes \cite{Li2025realsee3d_data} and evaluate both 8-panorama and 12-panorama input settings.

\subsubsection{Preprocessing}

The training preprocessing of the 2D generator follows DreamHome-Pano \cite{chen2026dreamhomepanodesignawareconflictfreepanoramic} in decomposing furnished panoramas into geometric and appearance conditions. We employ a defurnishing module based on Nano Banana 2 \cite{gemini_pro_3} and a fine-tuned Qwen-Image model \cite{wu2025qwenimagetechnicalreport} to obtain shell-like empty-room images. SAM \cite{kirillov2023segment} and MoGe-2 \cite{wang2025moge2} are then used to produce semantic segmentation maps and normal maps. The visual memory condition is rendered by our trained LRM from nearby panorama observations. For room grouping, 3D-FRONT labels are obtained from the relation between camera poses and wall meshes. RealSee3D room groups are coarsely annotated by estimating depth with DAP \cite{lin2026depthanypanoramas} and checking mutual visibility between cameras.

\subsubsection{Training Implementation}

The panoramic LRM is trained for seven days on 64 NVIDIA H200 GPUs. During training, it dynamically supports 1 to 24 input panoramas at a resolution of \(1024\times512\), with a global batch size of 256. The 2D panorama generator is trained with LoRA for four days on 8 NVIDIA H200 GPUs, with a global batch size of 16.

\subsubsection{Metrics}

We report HPSv3 \cite{Ma_2025_ICCV} as an aesthetic score for single-node visual quality. HPSv3 is a human-preference-aligned scoring model that has been shown to correlate well with human judgments of image aesthetics. We use image-image CLIP score \cite{hessel-etal-2021-clipscore} for style consistency with the reference image. For cross-node consistency, we compute Overlap PSNR (PSNR$_{\text{ov}}$): using the floorplan shell, we sample a set of 3D surface regions, mainly walls, wall-mounted decorations, and floors, project them into all panorama images that can observe them according to the camera poses, and compare the corresponding pixels (see supplementary material for details). We report PSNR, SSIM, and LPIPS \cite{zhang2018unreasonable} for whole-house LRM reconstruction quality. Higher is better for HPSv3, CLIP score, PSNR, and SSIM, while lower is better for LPIPS.

\begin{figure*}[t]
  \centering
  \includegraphics[width=\textwidth]{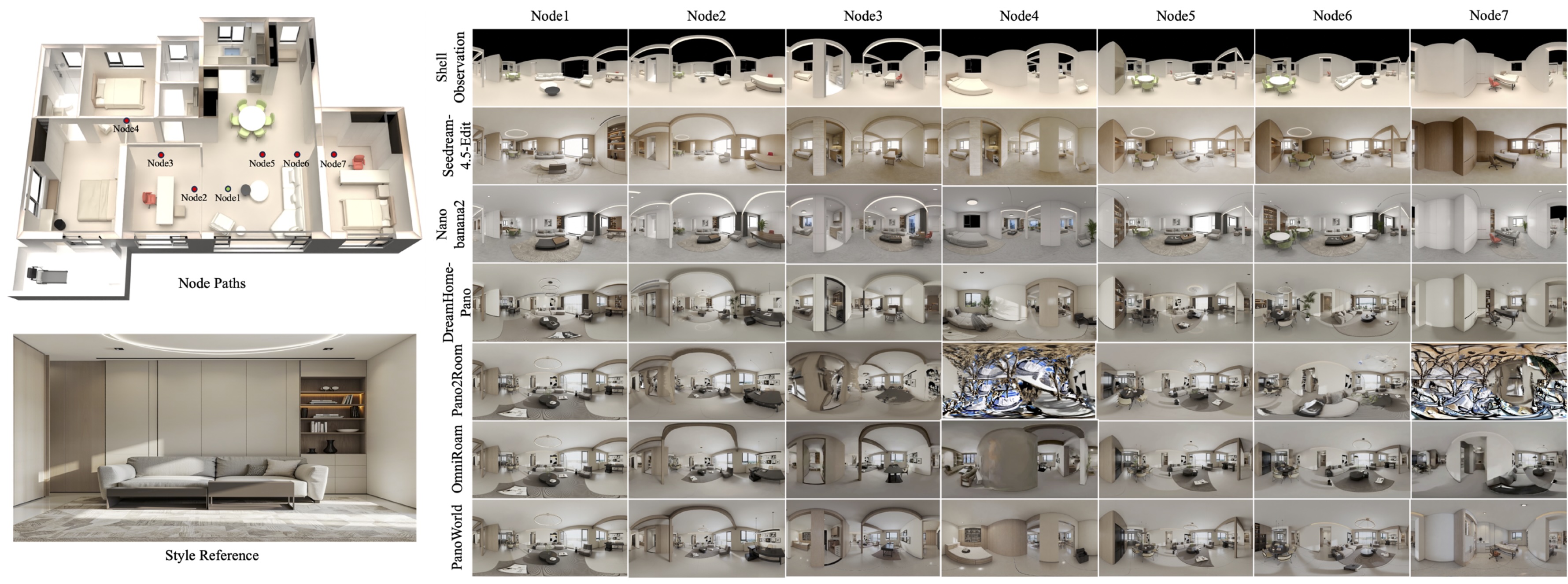}
  \caption{\textbf{Qualitative comparison on whole-house panorama synthesis.} We compare PanoWorld with representative adapted baselines on multi-node panorama generation.}
  \label{fig:qualitative_results}
\end{figure*}
\begin{figure*}[t]
  \centering
  \includegraphics[width=\textwidth]{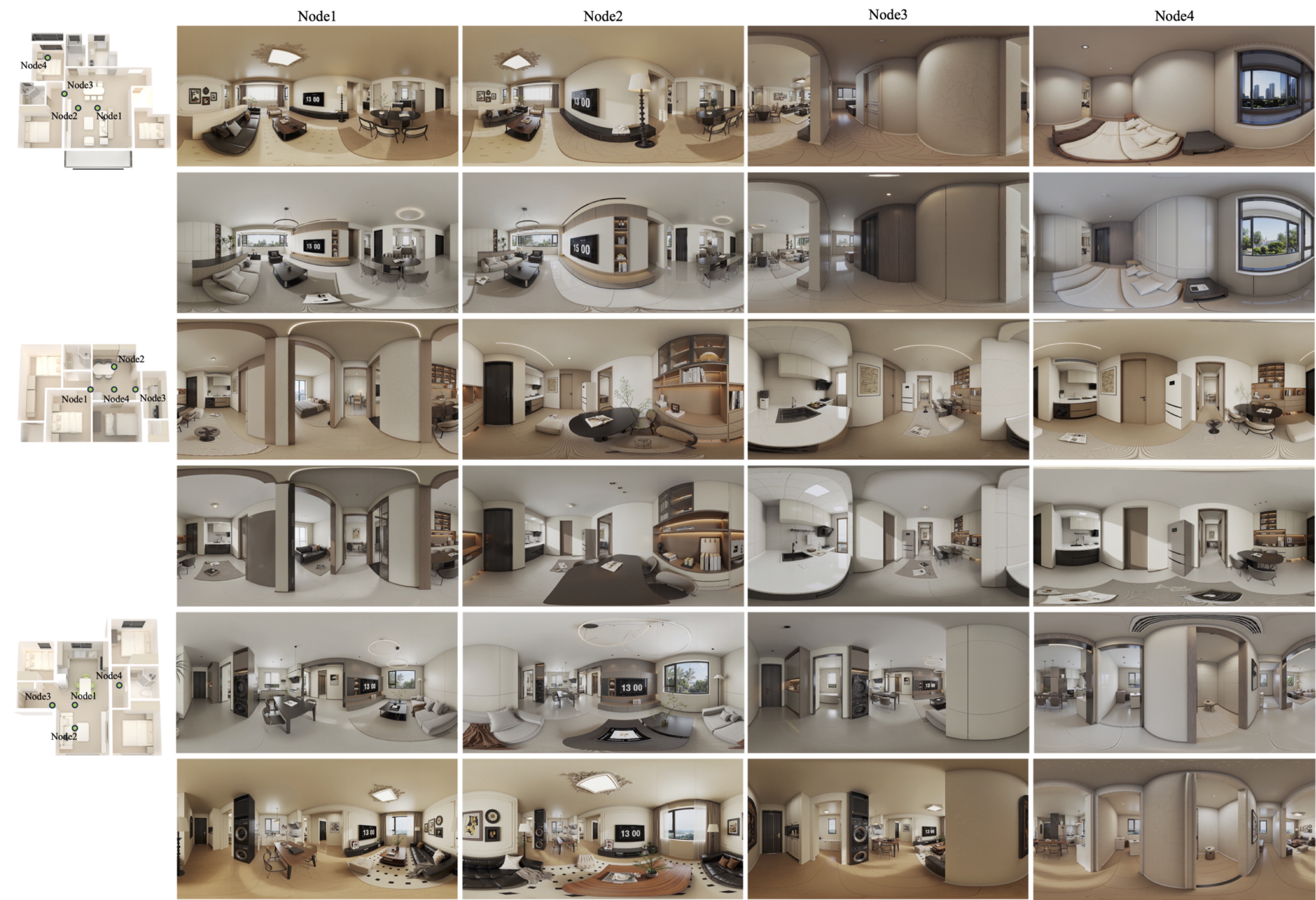}
  \caption{\textbf{PanoWorld qualitative results under different target styles.} PanoWorld preserves cross-room geometry and material identity while generating furnished panoramas under different target styles.}
  \label{fig:more_qualitative_results}
\end{figure*}
\subsection{Quantitative Results}

\subsubsection{Panorama Synthesis}

Since there is no existing academic task that exactly matches whole-house multi-node panorama synthesis, we adapt several representative methods to our setting for a relatively fair comparison. DreamHome-Pano \cite{chen2026dreamhomepanodesignawareconflictfreepanoramic} is a panorama generation model controlled by style and geometry, but it does not include an explicit multi-node consistency module. Pano2room \cite{pu2024pano2room} is adapted as a room-level panorama baseline without persistent whole-house memory. Nano Banana 2 \cite{gemini_pro_3} and Seedream-4.5-Edit \cite{chen2025seedream} are strong multimodal image editing models; we use text and image conditions to generate panoramas at target nodes. OmniRoam \cite{omniroam2026} is a panoramic video generation model adapted through progressive path-wise generation. The prompt, input formatting, and baseline adaptation protocols are described in the supplementary material. Table~\ref{tab:panorama_quant} compares these methods with PanoWorld on the released floorplan benchmark. The comparison covers both per-node perceptual quality and cross-node consistency. This separation is important because a method can produce attractive single panoramas while still drifting across nearby nodes. 

As shown in Table~\ref{tab:panorama_quant}, PanoWorld demonstrates a significant advantage in multi-node spatial consistency. It achieves an Overlap PSNR of 22.1365, outperforming the second-best method, OmniRoam, by a substantial margin of 5.75 dB. Regarding single-node aesthetic quality, most evaluated methods achieve an HPSv3 score around 7 to 8, with the notable exception of Pano2room. Nano Banana 2 obtains the highest single-node quality (9.5483) and style consistency (0.7940). Meanwhile, PanoWorld yields a competitive per-node visual quality (HPSv3 of 7.9564) while substantially reducing cross-node drift. This demonstrates that our method mitigates the geometry and material hallucinations prevalent in pure 2D generators, prioritizing structural coherence across the whole-house tour without compromising single-view photorealism.

\begin{figure*}[t]
  \centering
  \includegraphics[width=\textwidth]{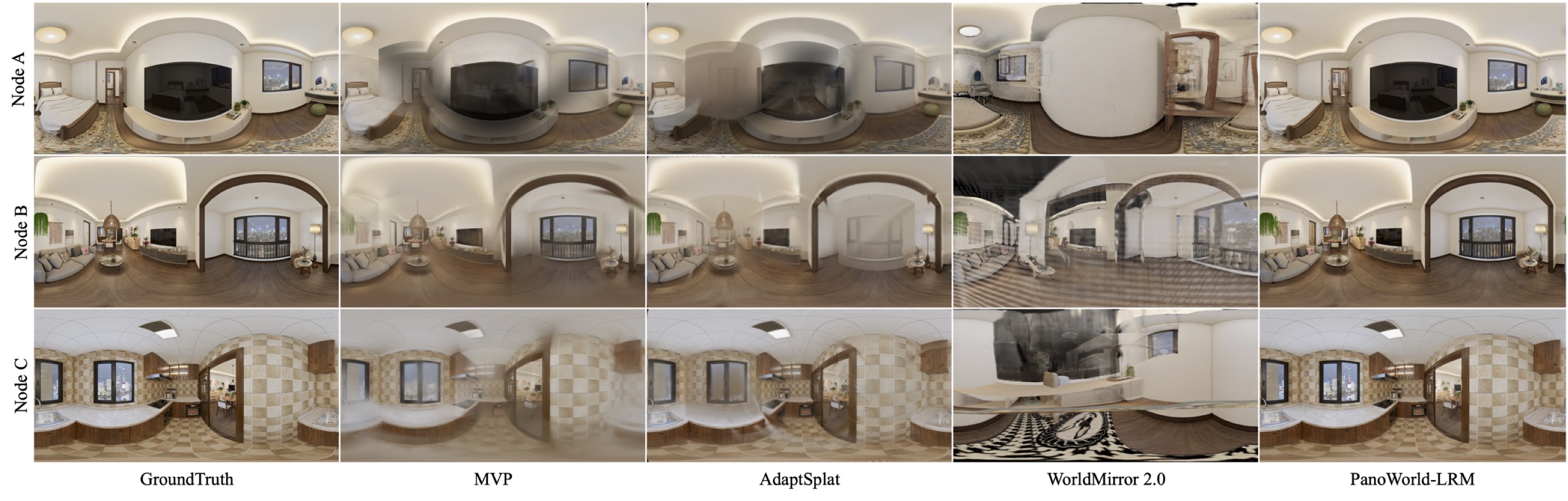}
  \caption{\textbf{Whole-house LRM reconstruction visualization.} The comparison shows room-level panorama renderings for different reconstruction methods.}
  \label{fig:lrm_visualization}
\end{figure*}

\subsubsection{Whole-House LRM Reconstruction}

Table~\ref{tab:lrm_reconstruction} evaluates the reconstruction quality of the panoramic LRM on 50 held-out RealSee3D scenes \cite{Li2025realsee3d_data}. We compare against MVP \cite{kang2025multi}, Adapt-Splat \cite{xing2026adaptsplatadaptingvisionfoundation}, and WorldMirror 2.0 \cite{hyworld22026} under 8-panorama and 12-panorama input settings. PanoWorld obtains the best reconstruction quality in both input settings, demonstrating its advantage in metric-scale multi-room whole-house reconstruction. The 12-panorama setting is slightly lower than the 8-panorama setting for PanoWorld because the additional viewpoints cover a larger spatial extent and introduce more cross-room visibility changes, making global multi-room fusion more challenging rather than simply providing redundant observations.

\subsection{Qualitative Results}

\subsubsection{Panorama Synthesis}

Figure~\ref{fig:qualitative_results} compares PanoWorld with representative adapted baselines for whole-house panorama synthesis. Figure~\ref{fig:more_qualitative_results} further presents PanoWorld results on additional floorplans and target styles. The examples highlight how PanoWorld turns shell geometry and cache-rendered visual memory into furnished panoramas while preserving alignment around doorways, corridors, living-dining connections, and cross-room views. The paired views show that geometry and material identity remain consistent when the same region is observed from different rooms.

\subsubsection{Whole-House LRM Reconstruction}

Figure~\ref{fig:lrm_visualization} visualizes representative panorama renderings from whole-house LRM reconstructions. Each column corresponds to one reconstruction method, and each row shows a selected room-level viewpoint. PanoWorld preserves sharper local textures and more coherent wall-door geometry across rooms, whereas competing methods exhibit blur, structural drift, or cross-room feature interference under multi-room inputs.

\subsection{Ablation Study}
\subsubsection{2D Generator Ablation}

Table~\ref{tab:ablation_2d_generator} studies the conditions used by the 2D generator. Removing the 3D cache tests whether nearby panorama conditioning alone can maintain spatial memory. Removing the nearby panorama tests whether cache rendering alone carries enough local appearance context. Removing Panoramic Position Encoding tests whether the generator can maintain equirectangular wraparound continuity without the Plucker extrinsics-only rays and CPRoPE. The results show that visual memory and nearby-view conditioning mainly improve cross-node consistency. Removing CPRoPE has little effect on HPSv3 but reduces PSNR$_{\text{ov}}$, indicating that its main role is preserving panorama-boundary continuity and cross-node geometric alignment rather than improving single-image aesthetics.

\subsubsection{LRM Ablation}

Table~\ref{tab:ablation_lrm} studies the reconstruction module. We remove CPRoPE and replace Room-Aware Group Attention (RAGA) with standard global attention to isolate the effects of panorama-aware spatial encoding and topology-aware cross-room feature aggregation. Removing RAGA causes the largest degradation, indicating that topology-aware attention is critical for multi-room reconstruction.

\begin{table}[t]
  \centering
  \caption{\textbf{Ablation study on the 2D panorama generator.} The table isolates the contribution of visual memory (VM), nearby-view conditioning (NV), and CPRoPE.}
  \label{tab:ablation_2d_generator}
  \begin{tabular}{lccc}
    \hline
    Variant & HPSv3 $\uparrow$ & CLIP-I Style $\uparrow$ & PSNR$_{\text{ov}}$ $\uparrow$ \\
    \hline
    Full PanoWorld & 7.9564 & 0.7577 & \textbf{22.1365} \\
    w/o VM & 7.8173 & \textbf{0.7591} & 18.8374 \\
    w/o NV & 7.8824 & 0.7381 & 19.0182 \\
    w/o CPRoPE & \textbf{7.9712} & 0.7586 & 20.6109 \\
    \hline
  \end{tabular}
\end{table}

\begin{table}[t]
  \centering
  \caption{\textbf{Ablation study on the panoramic LRM.} We evaluate CPRoPE and Room-Aware Group Attention (RAGA).}
  \label{tab:ablation_lrm}
  \begin{tabular}{lccc}
    \hline
    Variant & PSNR $\uparrow$ & SSIM $\uparrow$ & LPIPS $\downarrow$ \\
    \hline
    Full panoramic LRM & \textbf{29.2361} & \textbf{0.8880} & \textbf{0.2225} \\
    w/o CPRoPE & 28.1739 & 0.8622 & 0.2409 \\
    w/o RAGA & 21.7084 & 0.8216 & 0.2916 \\
    \hline
  \end{tabular}
\end{table}

\section{Discussion and Limitations}
\label{sec:discussion}

PanoWorld matches the node-based form of real indoor panorama tours while combining the texture fidelity of 2D generation with the spatial discipline of a renderable 3D memory. It also supports rapid global restyling because the shell geometry and cache-rendered memory are separated from the final appearance generator. Its limitations mainly come from imperfect geometry and sparse observation. Errors in the floorplan-to-shell process, missing doorway topology, or overly large spacing between panorama nodes can weaken cache guidance. Dynamic objects, mirrors, transparent materials, and heavy furniture occlusions remain challenging. Future work may jointly optimize shell estimation and generation, introduce object-level editable semantics, and improve interactive restyling.

\section{Conclusion}
\label{sec:conclusion}

We presented PanoWorld, a generative spatial world model for consistent whole-house panorama synthesis. By combining node-based autoregressive generation, a room-aware panoramic LRM, topology-aware progressive 3DGS caching, and decoupled geometry-appearance guidance, PanoWorld aims to generate high-fidelity furnished panoramas while preserving cross-node layout and material consistency across multi-room indoor tours.
{
    \small

\begin{thebibliography}{46}
\providecommand{\natexlab}[1]{#1}
\providecommand{\url}[1]{\texttt{#1}}
\expandafter\ifx\csname urlstyle\endcsname\relax
  \providecommand{\doi}[1]{doi: #1}\else
  \providecommand{\doi}{doi: \begingroup \urlstyle{rm}\Url}\fi

\bibitem[Charatan et~al.(2024)Charatan, Li, Sun, Luiten, Wetzstein, and
  Smith]{charatan2024pixelsplat}
David Charatan, Sizhe Li, Andrea Sun, Jonathon Luiten, Gordon Wetzstein, and
  Leonidas Smith.
\newblock pixelsplat: 3d gaussian splats from image pairs for scalable
  generalizable 3d reconstruction.
\newblock In \emph{Proceedings of the IEEE/CVF Conference on Computer Vision
  and Pattern Recognition (CVPR)}, pages 25828--25838, 2024.

\bibitem[Chen et~al.(2026)Chen, Hu, Liu, Li, and
  Yang]{chen2026dreamhomepanodesignawareconflictfreepanoramic}
Lulu Chen, Yijiang Hu, Yuanqing Liu, Yulong Li, and Yue Yang.
\newblock Dreamhome-pano: Design-aware and conflict-free panoramic interior
  generation, 2026.

\bibitem[Chen et~al.(2022)Chen, Wang, and Liu]{chen2022text2light}
Zhaoxi Chen, Guangcong Wang, and Ziwei Liu.
\newblock Text2light: Zero-shot text-driven hdr panorama generation.
\newblock \emph{ACM Transactions on Graphics (TOG)}, 41\penalty0 (6):\penalty0
  1--16, 2022.

\bibitem[Dhamo et~al.(2021)Dhamo, Bobrovsky, Navab, and
  Tombari]{dhamo2021graph}
Helisa Dhamo, Fabian Bobrovsky, Nassir Navab, and Federico Tombari.
\newblock Graph-to-3d: End-to-end generation and manipulation of 3d scenes
  using scene graphs.
\newblock In \emph{Proceedings of the IEEE/CVF International Conference on
  Computer Vision (ICCV)}, pages 16352--16361, 2021.

\bibitem[Feng et~al.(2023)Feng, Liu, Cui, and
  Xie]{feng2023diffusion360seamless360degree}
Mengyang Feng, Jinlin Liu, Miaomiao Cui, and Xuansong Xie.
\newblock Diffusion360: Seamless 360 degree panoramic image generation based on
  diffusion models, 2023.

\bibitem[Fridman et~al.(2023)Fridman, Carmeli, Dekel, and
  Michaeli]{fridman2023scenescape}
Rafail Fridman, Amit Carmeli, Tali Dekel, and Tomer Michaeli.
\newblock Scenescape: Text-driven consistent scene generation.
\newblock In \emph{SIGGRAPH Asia 2023 Conference Papers}, pages 1--10, 2023.

\bibitem[Fu et~al.(2021)Fu, Cai, Gao, Zhang, Li, Xun, Sun, Jia, Zhao, and
  Zhang]{fu20213dfront3dfurnishedrooms}
Huan Fu, Bowen Cai, Lin Gao, Lingxiao Zhang, Jiaming Wang~Cao Li, Zengqi Xun,
  Chengyue Sun, Rongfei Jia, Binqiang Zhao, and Hao Zhang.
\newblock 3d-front: 3d furnished rooms with layouts and semantics, 2021.

\bibitem[Google(2025)]{gemini_pro_3}
Google.
\newblock Nano banana pro, 2025.

\bibitem[He et~al.(2024)He, Wang, Kuang, Zhao, Wang, Chen, Luan, M{\"u}ller,
  Wang, Shen, et~al.]{he2024gslrm}
Zhenxing He, Zhisheng Wang, Yuhui Kuang, Min Zhao, Menglei Wang, Hao Chen,
  Fujun Luan, Thomas M{\"u}ller, Jiaqi Wang, Chunhua Shen, et~al.
\newblock Gs-lrm: Large reconstruction model for 3d gaussian splatting.
\newblock In \emph{European Conference on Computer Vision (ECCV)}. Springer,
  2024.

\bibitem[Hessel et~al.(2021)Hessel, Holtzman, Forbes, Le~Bras, and
  Choi]{hessel-etal-2021-clipscore}
Jack Hessel, Ari Holtzman, Maxwell Forbes, Ronan Le~Bras, and Yejin Choi.
\newblock {CLIPS}core: A reference-free evaluation metric for image captioning.
\newblock In \emph{Proceedings of the 2021 Conference on Empirical Methods in
  Natural Language Processing}, pages 7514--7528, Online and Punta Cana,
  Dominican Republic, 2021. Association for Computational Linguistics.

\bibitem[H{\"o}llein et~al.(2023)H{\"o}llein, Cao, Owens, Johnson, and
  Nie{\ss}ner]{hollein2023text2room}
Lukas H{\"o}llein, Ang Cao, Andrew Owens, Justin Johnson, and Matthias
  Nie{\ss}ner.
\newblock Text2room: Extracting textured 3d meshes from 2d text-to-image
  models.
\newblock In \emph{Proceedings of the IEEE/CVF International Conference on
  Computer Vision (ICCV)}, pages 7909--7920, 2023.

\bibitem[Hong et~al.(2024)Hong, Zhang, Gu, Bi, Zhou, Liu, Liu, Sunkavalli, Bui,
  and Tan]{Hong2024LRM}
Yicong Hong, Kai Zhang, Jiuxiang Gu, Sai Bi, Yang Zhou, Difan Liu, Feng Liu,
  Kalyan Sunkavalli, Trung Bui, and Hao Tan.
\newblock Lrm: Large reconstruction model for single image to 3d.
\newblock In \emph{ICLR}, 2024.

\bibitem[Hu et~al.(2024)Hu, Arroyo, Debats, Manhardt, Carlone, and
  Tombari]{Hu2024MiDiffusion}
Siyi Hu, Diego~Martin Arroyo, Stephanie Debats, Fabian Manhardt, Luca Carlone,
  and Federico Tombari.
\newblock Mixed diffusion for 3d indoor scene synthesis.
\newblock \emph{arXiv preprint arXiv:2405.21066}, 2024.

\bibitem[HY-World(2026)]{hyworld22026}
Team HY-World.
\newblock Hy-world 2.0: A multi-modal world model for reconstructing,
  generating, and simulating 3d worlds.
\newblock \emph{arXiv preprint}, 2026.

\bibitem[Jia et~al.(2026)Jia, Li, and Shi]{jia2026gaussianoncecontrollable3d}
Jinrang Jia, Zhenjia Li, and Yifeng Shi.
\newblock You only gaussian once: Controllable 3d gaussian splatting for
  ultra-densely sampled scenes, 2026.

\bibitem[Kang et~al.(2025)Kang, Yang, Nam, Lee, Kim, and Park]{kang2025multi}
Gyeongjin Kang, Seungkwon Yang, Seungtae Nam, Younggeun Lee, Jungwoo Kim, and
  Eunbyung Park.
\newblock Multi-view pyramid transformer: Look coarser to see broader.
\newblock \emph{arXiv preprint arXiv:2512.07806}, 2025.

\bibitem[Kerbl et~al.(2023)Kerbl, Kopanas, Leimk{\"u}hler, and
  Drettakis]{kerbl3Dgaussians}
Bernhard Kerbl, Georgios Kopanas, Thomas Leimk{\"u}hler, and George Drettakis.
\newblock 3d gaussian splatting for real-time radiance field rendering.
\newblock \emph{ACM Transactions on Graphics}, 42\penalty0 (4), 2023.

\bibitem[Kirillov et~al.(2023)Kirillov, Mintun, Ravi, Mao, Rolland, Gustafson,
  Xiao, Whitehead, Berg, Lo, Doll{\'a}r, and Girshick]{kirillov2023segment}
Alexander Kirillov, Eric Mintun, Nikhila Ravi, Hanzi Mao, Chloe Rolland, Laura
  Gustafson, Tete Xiao, Spencer Whitehead, Alexander~C Berg, Wan-Yen Lo, Piotr
  Doll{\'a}r, and Ross Girshick.
\newblock Segment anything.
\newblock In \emph{Proceedings of the IEEE/CVF International Conference on
  Computer Vision}, pages 4015--4026, 2023.

\bibitem[Li and Bansal(2023)]{Li2023PanoGen}
Jialu Li and Mohit Bansal.
\newblock Panogen: Text-conditioned panoramic environment generation for
  vision-and-language navigation.
\newblock In \emph{NeurIPS}, 2023.

\bibitem[Li et~al.(2024{\natexlab{a}})Li, Tan, Zhang, Xu, Luan, Xu, Hong,
  Sunkavalli, Shakhnarovich, and Bi]{Li2024Instant3D}
Jiahao Li, Hao Tan, Kai Zhang, Zexiang Xu, Fujun Luan, Yinghao Xu, Yicong Hong,
  Kalyan Sunkavalli, Greg Shakhnarovich, and Sai Bi.
\newblock Instant3d: Fast text-to-3d with sparse-view generation and large
  reconstruction model.
\newblock In \emph{ICLR}, 2024{\natexlab{a}}.

\bibitem[Li et~al.(2025{\natexlab{a}})Li, Wu, Li, Wang, Rao, Zhou, Pan, and
  Hui]{Li2025realsee3d_data}
Linyuan Li, Yan Wu, Xi Li, Lingli Wang, Tong Rao, Jie Zhou, Cihui Pan, and
  Xinchen Hui.
\newblock Realsee3d: A large-scale multi-view rgb-d dataset of indoor scenes
  (version 1.0), 2025{\natexlab{a}}.

\bibitem[Li et~al.(2024{\natexlab{b}})Li, Long, Liang, Li, Liu, Li, Wang, Qi,
  Xue, Luo, Liu, and Guo]{Li2024MLRM}
Mengfei Li, Xiaoxiao Long, Yixun Liang, Weiyu Li, Yuan Liu, Peng Li, Yatian
  Wang, Xingqun Qi, Wei Xue, Wenhan Luo, Qifeng Liu, and Yike Guo.
\newblock M-lrm: Multi-view large reconstruction model.
\newblock \emph{arXiv preprint arXiv:2406.07648}, 2024{\natexlab{b}}.

\bibitem[Li et~al.(2025{\natexlab{b}})Li, Yi, Liu, Gao, Ma, and
  Kanazawa]{li2025cameras}
Ruilong Li, Brent Yi, Junchen Liu, Hang Gao, Yi Ma, and Angjoo Kanazawa.
\newblock Cameras as relative positional encoding.
\newblock In \emph{Advances in Neural Information Processing Systems},
  2025{\natexlab{b}}.

\bibitem[Lin et~al.(2026)Lin, Song, Zhang, Lu, Li, Du, Yang, Nguyen, and
  Qi]{lin2026depthanypanoramas}
Xin Lin, Meixi Song, Dizhe Zhang, Wenxuan Lu, Haodong Li, Bo Du, Ming-Hsuan
  Yang, Truong Nguyen, and Lu Qi.
\newblock Depth any panoramas: A foundation model for panoramic depth
  estimation.
\newblock In \emph{Proceedings of the IEEE/CVF Conference on Computer Vision
  and Pattern Recognition}, 2026.

\bibitem[Liu et~al.(2026)Liu, Lin, Li, Yang, Wang, Sunkavalli, Hold-Geoffroy,
  Tan, Zhang, Xie, Shi, and Hu]{omniroam2026}
Yuheng Liu, Xin Lin, Xinke Li, Baihan Yang, Chen Wang, Kalyan Sunkavalli,
  Yannick Hold-Geoffroy, Hao Tan, Kai Zhang, Xiaohui Xie, Zifan Shi, and Yiwei
  Hu.
\newblock Omniroam: World wandering via long-horizon panoramic video
  generation.
\newblock \emph{SIGGRAPH}, 2026.

\bibitem[Ma et~al.(2025)Ma, Wu, Sun, and Li]{Ma_2025_ICCV}
Yuhang Ma, Xiaoshi Wu, Keqiang Sun, and Hongsheng Li.
\newblock Hpsv3: Towards wide-spectrum human preference score.
\newblock In \emph{Proceedings of the IEEE/CVF International Conference on
  Computer Vision (ICCV)}, pages 15086--15095, 2025.

\bibitem[Mildenhall et~al.(2020)Mildenhall, Srinivasan, Tancik, Barron,
  Ramamoorthi, and Ng]{mildenhall2020nerf}
Ben Mildenhall, Pratul~P. Srinivasan, Matthew Tancik, Jonathan~T. Barron, Ravi
  Ramamoorthi, and Ren Ng.
\newblock Nerf: Representing scenes as neural radiance fields for view
  synthesis.
\newblock In \emph{European Conference on Computer Vision (ECCV)}, pages
  405--421. Springer, 2020.

\bibitem[Ni et~al.(2025)Ni, Zhang, Zhang, and Zhang]{Ni_2025_ICCV}
Jinhong Ni, Chang-Bin Zhang, Qiang Zhang, and Jing Zhang.
\newblock What makes for text to 360-degree panorama generation with stable
  diffusion?
\newblock In \emph{Proceedings of the IEEE/CVF International Conference on
  Computer Vision (ICCV)}, 2025.

\bibitem[Paschalidou et~al.(2021)Paschalidou, Kar, Shugrina, Kreis, Geiger, and
  Fidler]{Paschalidou2021ATISS}
Despoina Paschalidou, Amlan Kar, Maria Shugrina, Karsten Kreis, Andreas Geiger,
  and Sanja Fidler.
\newblock Atiss: Autoregressive transformers for indoor scene synthesis.
\newblock In \emph{NeurIPS}, 2021.

\bibitem[Pu et~al.(2024)Pu, Zhao, and Lian]{pu2024pano2room}
Guo Pu, Yiming Zhao, and Zhouhui Lian.
\newblock Pano2room: Novel view synthesis from a single indoor panorama.
\newblock In \emph{SIGGRAPH Asia 2024 Conference Papers}, New York, NY, USA,
  2024. Association for Computing Machinery.

\bibitem[Rombach et~al.(2022)Rombach, Blattmann, Lorenz, Esser, and
  Ommer]{rombach2022high}
Robin Rombach, Andreas Blattmann, Dominik Lorenz, Patrick Esser, and Bj{\"o}rn
  Ommer.
\newblock High-resolution image synthesis with latent diffusion models.
\newblock In \emph{Proceedings of the IEEE/CVF Conference on Computer Vision
  and Pattern Recognition (CVPR)}, 2022.

\bibitem[Shabani et~al.(2023)Shabani, Hosseini, and
  Furukawa]{shabani2023housediffusion}
Amin Shabani, Sepideh Hosseini, and Yasutaka Furukawa.
\newblock Housediffusion: Vector floorplan generation via a diffusion model.
\newblock In \emph{Proceedings of the IEEE/CVF Conference on Computer Vision
  and Pattern Recognition (CVPR)}, pages 5466--5475, 2023.

\bibitem[Tang et~al.(2024{\natexlab{a}})Tang, Chen, Chen, Wang, Zeng, and
  Liu]{Tang2024LGM}
Jiaxiang Tang, Zhaoxi Chen, Xiaokang Chen, Tengfei Wang, Gang Zeng, and Ziwei
  Liu.
\newblock Lgm: Large multi-view gaussian model for high-resolution 3d content
  creation.
\newblock In \emph{ECCV}, 2024{\natexlab{a}}.

\bibitem[Tang et~al.(2024{\natexlab{b}})Tang, Nie, Markhasin, Dai, Thies, and
  Nie{\ss}ner]{tang2024diffuscene}
Jiapeng Tang, Yinyu Nie, Lev Markhasin, Angela Dai, Justus Thies, and Matthias
  Nie{\ss}ner.
\newblock Diffuscene: Denoising diffusion probabilistic models for generative
  indoor scene synthesis.
\newblock In \emph{Proceedings of the IEEE/CVF Conference on Computer Vision
  and Pattern Recognition (CVPR)}, 2024{\natexlab{b}}.

\bibitem[Tang et~al.(2023)Tang, Zhang, Chen, Wang, and
  Furukawa]{Tang2023MVDiffusion}
Shitao Tang, Fuyang Zhang, Jiacheng Chen, Peng Wang, and Yasutaka Furukawa.
\newblock Mvdiffusion: Enabling holistic multi-view image generation with
  correspondence-aware diffusion.
\newblock In \emph{NeurIPS}, 2023.

\bibitem[{Team Seedream} et~al.(2025){Team Seedream}, Chen, Gao, Gong, Guo,
  Guo, et~al.]{chen2025seedream}
{Team Seedream}, Yunpeng Chen, Yu Gao, Lixue Gong, Meng Guo, Qiushan Guo,
  et~al.
\newblock Seedream 4.0: Toward next-generation multimodal image generation.
\newblock \emph{arXiv preprint arXiv:2509.20427}, 2025.

\bibitem[Tochilkin et~al.(2024)Tochilkin, Pankratz, Liu, Huang, Letts, Li,
  Liang, Laforte, Jampani, and Cao]{Tochilkin2024TripoSR}
Dmitry Tochilkin, David Pankratz, Zexiang Liu, Zixuan Huang, Adam Letts,
  Yangguang Li, Ding Liang, Christian Laforte, Varun Jampani, and Yan-Pei Cao.
\newblock Triposr: Fast 3d object reconstruction from a single image.
\newblock \emph{arXiv preprint arXiv:2403.02151}, 2024.

\bibitem[Vidanapathirana et~al.(2021)Vidanapathirana, Wu, Furukawa, Chang, and
  Savva]{vidanapathirana2021plan2scene}
Madhawa Vidanapathirana, Qirui Wu, Yasutaka Furukawa, Angel~X Chang, and
  Manolis Savva.
\newblock Plan2scene: Converting floorplans to 3d scenes.
\newblock In \emph{Proceedings of the IEEE/CVF Conference on Computer Vision
  and Pattern Recognition (CVPR)}, pages 10733--10742, 2021.

\bibitem[Wang et~al.(2025)Wang, Xu, Dong, Deng, Xiang, Lv, Sun, Tong, and
  Yang]{wang2025moge2}
Ruicheng Wang, Sicheng Xu, Yue Dong, Yu Deng, Jianfeng Xiang, Zelong Lv,
  Guangzhong Sun, Xin Tong, and Jiaolong Yang.
\newblock Moge-2: Accurate monocular geometry with metric scale and sharp
  details.
\newblock In \emph{Advances in Neural Information Processing Systems}, 2025.

\bibitem[Wang et~al.(2021)Wang, Yeh, Tang, Robbins, and
  Wang]{wang2021sceneformer}
Xin-Yang Wang, Yu-An Yeh, Che-Wei Tang, Anton Robbins, and Yu-Chiang~Frank
  Wang.
\newblock Sceneformer: Indoor scene generation with transformers.
\newblock In \emph{2021 International Conference on 3D Vision (3DV)}, pages
  106--115. IEEE, 2021.

\bibitem[Wu et~al.(2025)Wu, Li, Zhou, Lin, Gao, Yan, ming Yin, Bai, Xu, Chen,
  Chen, Tang, Zhang, Wang, Yang, Yu, Cheng, Liu, Li, Zhang, Meng, Wei, Ni,
  Chen, Cao, Peng, Qu, Wu, Wang, Yu, Wen, Feng, Xu, Wang, Zhang, Zhu, Wu, Cai,
  and Liu]{wu2025qwenimagetechnicalreport}
Chenfei Wu, Jiahao Li, Jingren Zhou, Junyang Lin, Kaiyuan Gao, Kun Yan, Sheng
  ming Yin, Shuai Bai, Xiao Xu, Yilei Chen, Yuxiang Chen, Zecheng Tang, Zekai
  Zhang, Zhengyi Wang, An Yang, Bowen Yu, Chen Cheng, Dayiheng Liu, Deqing Li,
  Hang Zhang, Hao Meng, Hu Wei, Jingyuan Ni, Kai Chen, Kuan Cao, Liang Peng,
  Lin Qu, Minggang Wu, Peng Wang, Shuting Yu, Tingkun Wen, Wensen Feng,
  Xiaoxiao Xu, Yi Wang, Yichang Zhang, Yongqiang Zhu, Yujia Wu, Yuxuan Cai, and
  Zenan Liu.
\newblock Qwen-image technical report, 2025.

\bibitem[Wu et~al.(2023)Wu, Zheng, and Cham]{Wu2023PanoDiffusion}
Tianhao Wu, Chuanxia Zheng, and Tat-Jen Cham.
\newblock Panodiffusion: 360-degree panorama outpainting via diffusion.
\newblock \emph{arXiv preprint arXiv:2307.03177}, 2023.

\bibitem[Xing et~al.(2026)Xing, Wang, and
  Shi]{xing2026adaptsplatadaptingvisionfoundation}
Mingwei Xing, Xinliang Wang, and Yifeng Shi.
\newblock Adaptsplat: Adapting vision foundation models for feed-forward 3d
  gaussian splatting, 2026.

\bibitem[Zhang et~al.(2024)Zhang, Wu, Gambardella, Huang, Phung, Ouyang, and
  Cai]{Zhang2024PanFusion}
Cheng Zhang, Qianyi Wu, Camilo~Cruz Gambardella, Xiaoshui Huang, Dinh Phung,
  Wanli Ouyang, and Jianfei Cai.
\newblock Taming stable diffusion for text to 360 degree panorama image
  generation.
\newblock In \emph{CVPR}, pages 6347--6357, 2024.

\bibitem[Zhang et~al.(2025)Zhang, Xu, Wu, Gambardella, Phung, and
  Cai]{zhang2024pansplat}
Cheng Zhang, Haofei Xu, Qianyi Wu, Camilo~Cruz Gambardella, Dinh Phung, and
  Jianfei Cai.
\newblock Pansplat: 4k panorama synthesis with feed-forward gaussian splatting.
\newblock In \emph{Proceedings of the IEEE/CVF Conference on Computer Vision
  and Pattern Recognition}, 2025.

\bibitem[Zhang et~al.(2018)Zhang, Isola, Efros, Shechtman, and
  Wang]{zhang2018unreasonable}
Richard Zhang, Phillip Isola, Alexei~A Efros, Eli Shechtman, and Oliver Wang.
\newblock The unreasonable effectiveness of deep features as a perceptual
  metric.
\newblock In \emph{Proceedings of the IEEE conference on computer vision and
  pattern recognition}, pages 586--595, 2018.

\end{thebibliography}

}

\clearpage
\setcounter{page}{1}
\maketitlesupplementary

\section{Visualization without Panoramic Position Encoding}

We include qualitative failure cases for the variant without Panoramic Position Encoding. Without circular horizontal encoding, the generator treats the left and right panorama boundaries as distant image regions rather than adjacent rays. This often produces inconsistent structures or textures across the seam, such as broken wall patterns, discontinuous furniture edges, or mismatched lighting at the panorama boundary.

\begin{figure}[t]
  \centering
  \includegraphics[width=\linewidth]{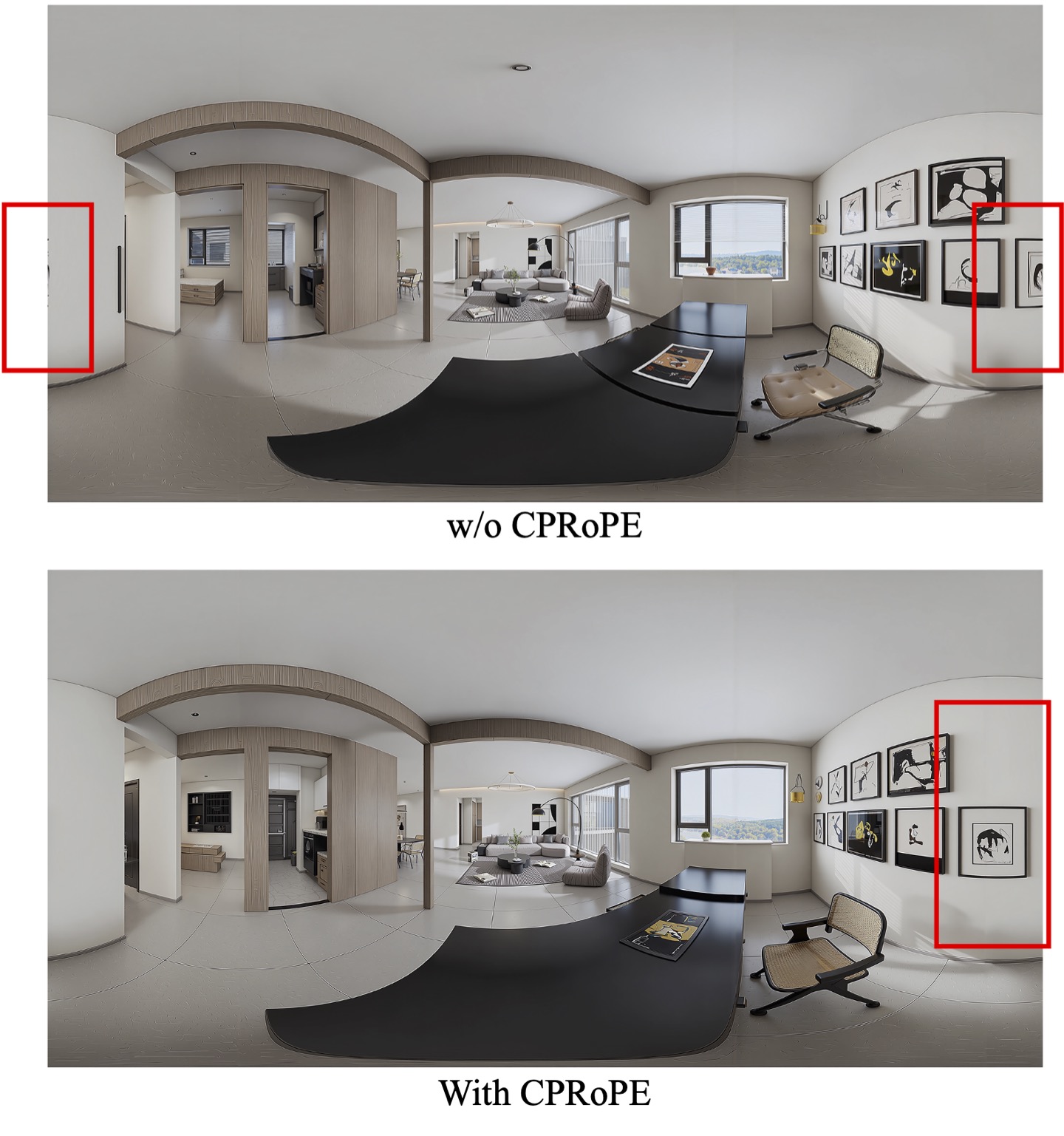}
  \caption{\textbf{Effect of panoramic position encoding.} Removing circular panoramic encoding causes left-right inconsistency and seam artifacts in generated panoramas.}
  \label{fig:appendix_no_pano_position}
\end{figure}

\section{Baseline Adaptation Details}
\label{app:baseline_adaptation}

\subsection{Pano2room}

Pano2room shares a broadly similar pipeline with our method, relying on monocular depth estimation to obtain a point cloud that is subsequently converted into a mesh, followed by iterative refinement through a render-then-estimate loop to progressively extend the scene to more distant regions. However, Pano2room decomposes the panoramic image into a set of perspective sub-images for depth estimation and then reprojects the results back into the panoramic coordinate system. This decomposition strategy substantially increases algorithmic complexity and compromises depth consistency across viewing directions. Furthermore, due to the absence of prior information such as placeholder images at target viewpoints, and because the inpainting backbone Stable Diffusion 2 \cite{rombach2022high} has not been fine-tuned for this task, the method tends to suffer from global scene collapse when confronted with large missing regions.

\subsection{Image-Editing Baseline Protocol}

Nano Banana 2 and Seedream-4.5-Edit are adapted as image-editing baselines. Since neither model maintains an explicit persistent 3D memory, each target panorama is generated independently from a geometry-control image and a style reference image. The following subsections describe the concrete model, prompt, and input simplification used for each baseline.

\subsection{Nano Banana 2}

For Nano Banana 2 \cite{gemini_pro_3}, we use the Gemini-3.1-flash-image-preview model. The first input image is the geometry-control image for the target panorama, and the second input image is the style reference. We use the following fixed text prompt:
\begin{quote}
\small
Generate a new image based on the spatial structure and furniture layout of the first image (control) and the style, shape, material, color, texture of the second image (style reference). Follow the following description: A modern minimalist living room featuring a clean, uncluttered aesthetic with a neutral color palette. The walls are finished in smooth white gypsum board with large panel design, creating a bright and expansive backdrop that enhances the sense of space. The ceiling maintains a simple flat design with subtle gypsum lines adding delicate architectural detail. The flooring consists of light grey micro cement with a smooth, flat finish that seamlessly integrates with the overall minimalist aesthetic. The room features a large modular sofa upholstered in light grey fabric, providing ample seating with its generous proportions and clean lines. In front of the sofa sits a rectangular combination coffee table in dark grey, constructed from wood and metal elements that complement the room's modern sensibility. A built-in open style bookshelf with wood color finish adds functional storage and visual interest, illuminated by integrated lighting that highlights its contents. The space is anchored by a large rectangular woven rug in light grey with geometric pattern that defines the seating area while adding subtle texture. Ceiling lighting includes minimalist round fixtures that provide overall illumination without visual clutter. Decorative elements are intentionally restrained, including rectangular throw pillows in coordinating light and dark grey tones, and a few carefully selected books and ceramic ornaments that add personality without disrupting the clean aesthetic. A large areca palm plant in the corner introduces natural greenery and softens the room's architectural lines. The overall atmosphere is one of calm sophistication, achieved through thoughtful proportions, restrained color palette, and intentional negative space that allows each element to breathe and be appreciated.
\end{quote}

\begin{figure}[t]
  \centering
  \includegraphics[width=\linewidth]{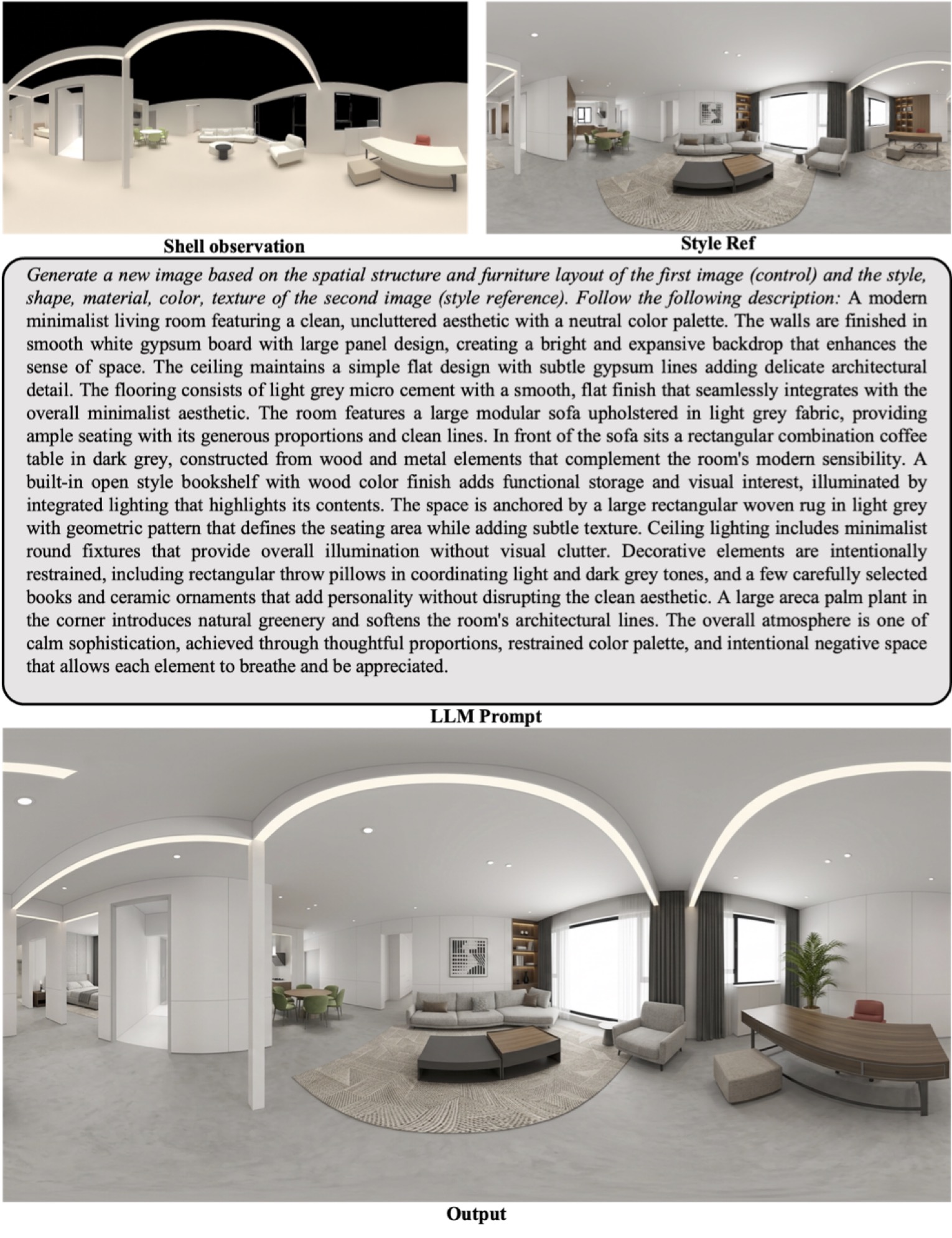}
  \caption{\textbf{Nano Banana 2 adaptation pipeline.} We use Gemini-3.1-flash-image-preview with a geometry-control image, a style reference, and a fixed descriptive prompt.}
  \label{fig:appendix_nanobanana_pipeline}
\end{figure}

\subsection{Seedream-4.5-Edit}

For Seedream-4.5-Edit, directly feeding the shell-rendered image and a complex text prompt often fails to produce a panorama that respects equirectangular distortion, because the model shows weaker understanding of the image and text conditions in this setting. We therefore simplify the protocol in two steps. First, we convert the shell image into a line drawing to emphasize the spatial structure. Second, we use a simplified fixed prompt for panoramic rendering:
\begin{quote}
\small
Please render this first panoramic line drawing into a panoramic rendering, keeping the spatial structure and furniture layout completely consistent with the first line drawing. Refer to the second image for style and furniture elements.
\end{quote}

\begin{figure}[t]
  \centering
  \includegraphics[width=\linewidth]{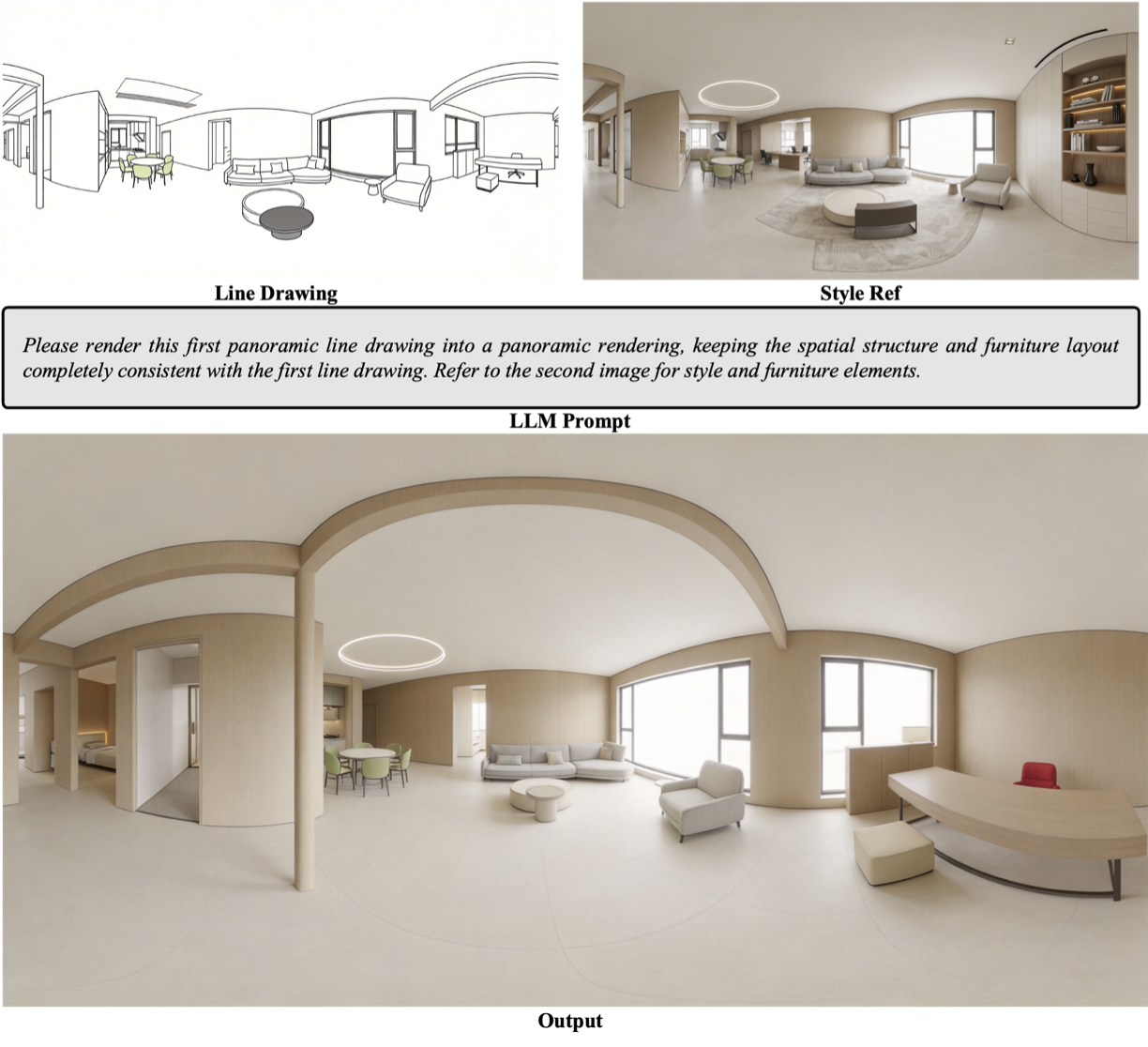}
  \caption{\textbf{Seedream-4.5-Edit adaptation pipeline.} We convert the shell image into a line drawing and use a simplified prompt to improve spatial-structure following.}
  \label{fig:appendix_seedream_pipeline}
\end{figure}

\subsection{OmniRoam}

We adapt OmniRoam \cite{omniroam2026} via progressive video generation. Starting from the first panorama node, we generate a short panoramic video segment toward the next navigation node along the planned path. The synthesized frame closest to the target node is then reprojected into an equirectangular panorama and used as the visual condition for the subsequent segment; repeating this process simulates a multi-node panorama tour. Estimating synthesized frame positions using the official scale reported by OmniRoam, 0.25m per latent step, yielded unsatisfactory results; through scale analysis, we empirically set the latent step to 0.1m to obtain reasonable position estimates. This protocol gives the video-based baseline access to local temporal continuity, but it still lacks PanoWorld's topology-aware 3DGS cache and room-aware reconstruction loop.

\section{Training Data Visualization}
\label{app:training_data_visualization}

\begin{figure*}[t]
  \centering
  \includegraphics[width=\textwidth]{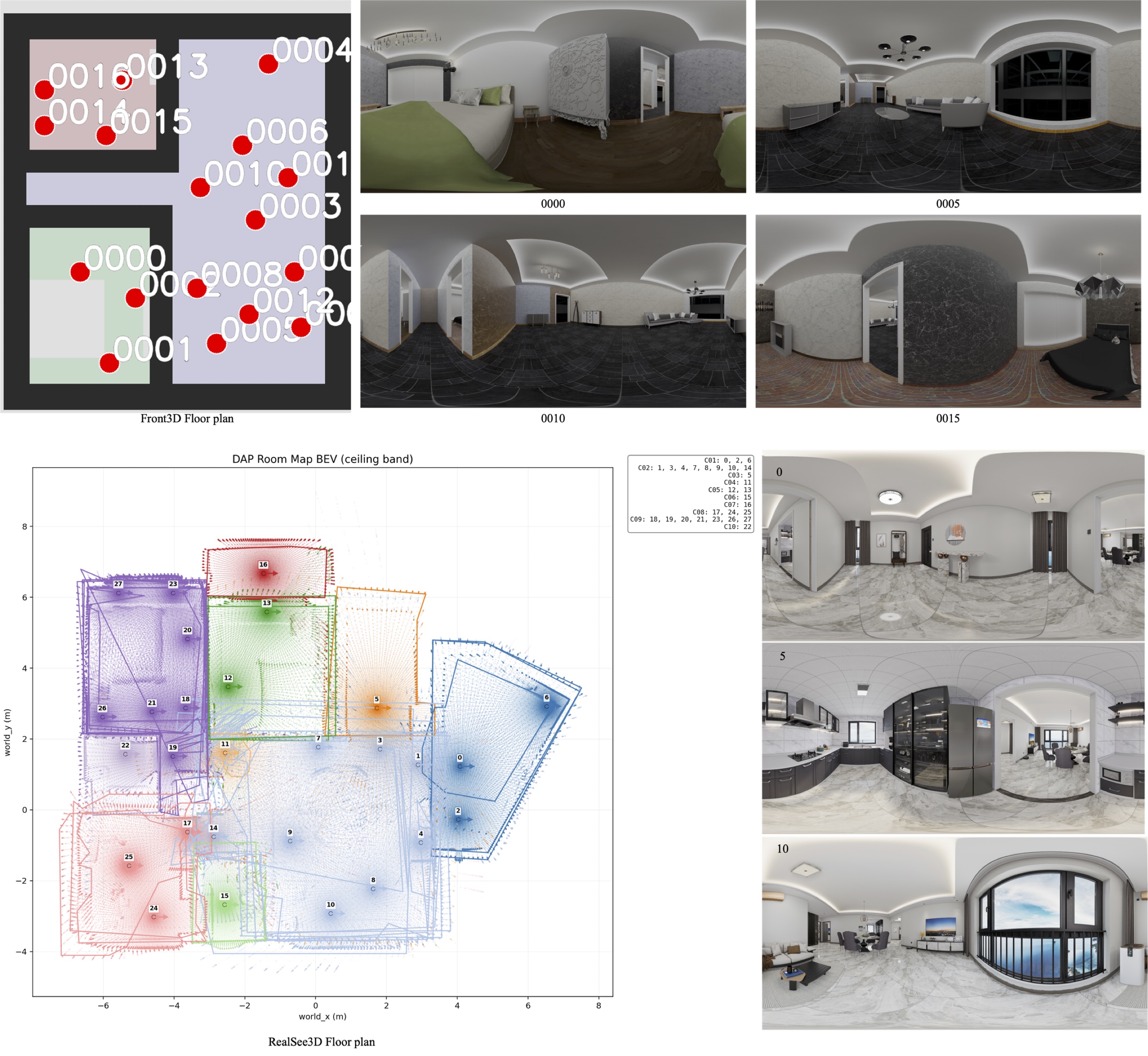}
  \caption{\textbf{Training data visualization.} Examples from 3D-FRONT and RealSee3D with panoramas, depth or shell-proxy images, and room-level BEV maps showing room partitions, doorways, sampled camera nodes, and local room groups.}
  \label{fig:appendix_training_data}
\end{figure*}

We further visualize the training data used by PanoWorld. Figure~\ref{fig:appendix_training_data} shows representative examples from 3D-FRONT \cite{fu20213dfront3dfurnishedrooms} and RealSee3D \cite{Li2025realsee3d_data}, including rendered panoramas, depth or shell-proxy images, and room-level BEV maps. The BEV maps illustrate the floorplan topology, room partitions, doorway connectivity, sampled camera nodes, and local room groups used to construct the training views. These visualizations clarify the difference between synthetic CAD-derived scenes and reconstructed real indoor scenes, and show how both sources support metric-scale multi-room panoramic training.

\section{Additional Experimental Details}
\label{app:experimental_details}

We include implementation details that are useful for reproducing the evaluation but too specific for the main paper, including panorama resolution, node sampling rules, and overlap-mask construction for cross-view PSNR.

\subsection{Cross-Node Consistency Evaluation}
\label{app:cross_node_consistency}

We evaluate cross-node consistency on manually selected co-visible surface regions defined on the 3D shell asset. For each scene, we first choose an initial panorama node and identify several visible \(1\mathrm{m}\times1\mathrm{m}\) evaluation patches on walls, floors, paintings, or other wall-mounted decorations. Each patch is sampled at a 1cm interval along its two surface axes, producing \(100\times100=10{,}000\) 3D points:
\[
\mathcal{P}_r=\{x_r+a\Delta e^u_r+b\Delta e^v_r\mid a,b\in\{0,\ldots,99\},\Delta=0.01\mathrm{m}\},
\]
where \(x_r\) is one corner of region \(r\), and \(e^u_r,e^v_r\) are orthonormal directions on the selected surface. Given camera extrinsics, each sampled point \(p\in\mathcal{P}_r\) is projected to the initial node and to an evaluated node by equirectangular projection, yielding corresponding pixels \(\pi_0(p)\) and \(\pi_t(p)\). We then compute the pixel MSE and PSNR over valid co-visible samples:
\[
\begin{aligned}
\mathrm{MSE}_{0,t,r}
&=\frac{1}{|\Omega_r|}\sum_{p\in\Omega_r}
\left\|I_0(\pi_0(p))-I_t(\pi_t(p))\right\|_2^2,\\
\mathrm{PSNR}_{0,t,r}
&=10\log_{10}\frac{255^2}{\mathrm{MSE}_{0,t,r}} .
\end{aligned}
\]
The final overlap PSNR averages over all selected regions and evaluated nodes. We focus on walls, floors, and paintings because these regions are usually planar, have unified shell geometry, and suffer little geometric self-occlusion. They are also sensitive to appearance drift: a weakly textured white wall in the initial view may become patterned wallpaper or acquire inconsistent material details in another node, making cross-node inconsistency clearly measurable.

\begin{figure*}[t]
  \centering
  \includegraphics[width=\textwidth]{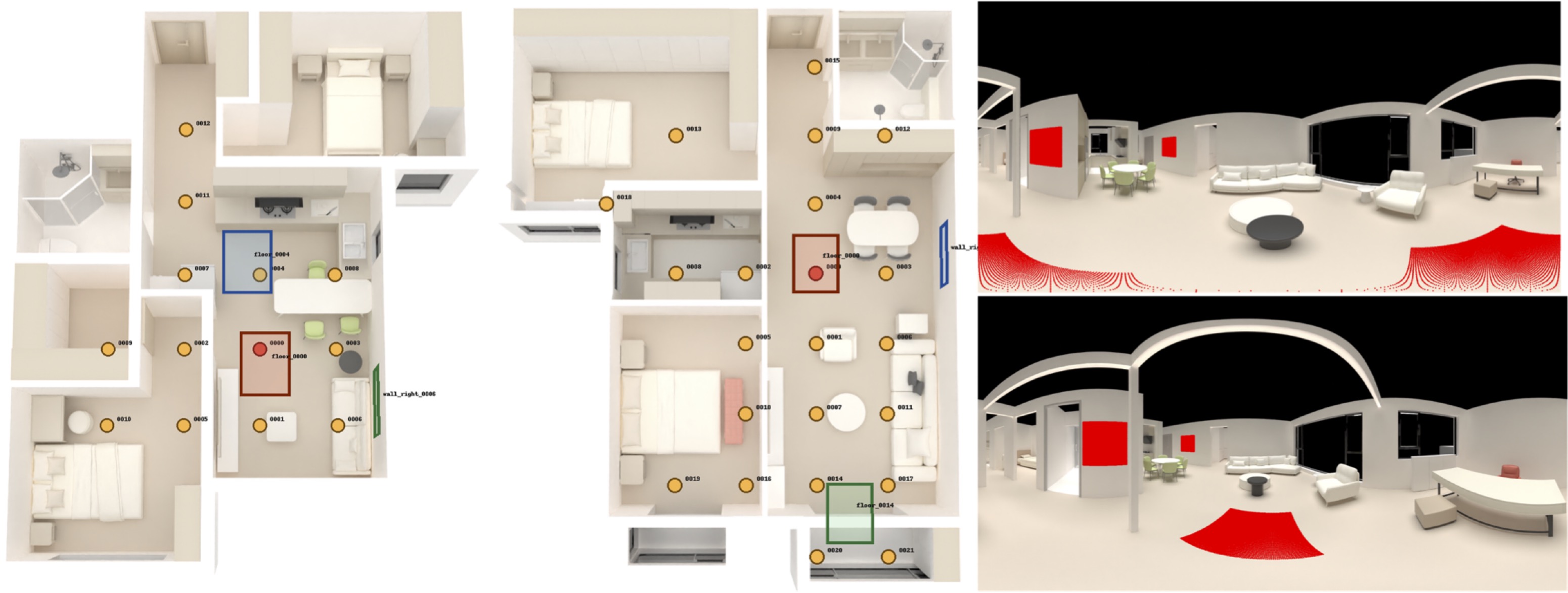}
  \caption{\textbf{Cross-node consistency evaluation regions.} We manually select co-visible \(1\mathrm{m}\times1\mathrm{m}\) regions on planar shell surfaces, densely sample 3D points, project them into multiple panorama nodes, and compute PSNR over corresponding pixels.}
  \label{fig:appendix_cross_node_regions}
\end{figure*}

\end{document}